\documentclass[runningheads]{llncs}

\usepackage{accv}

\usepackage{accvabbrv}

\usepackage{microtype}
\usepackage{graphicx}
\usepackage{booktabs}
\usepackage{multirow}
\usepackage[table]{xcolor}
\usepackage{float}
\usepackage{tcolorbox}
\usepackage{esvect}
\tcbuselibrary{breakable}

\usepackage[accsupp]{axessibility}  

\usepackage{hyperref}

\usepackage{orcidlink}

\definecolor{baselinebg}{RGB}{242,242,242}
\definecolor{bestragbg}{RGB}{220,238,255}
\usepackage{amsmath}
\usepackage{amssymb}
\usepackage{rotating}
\tcbuselibrary{breakable,skins}
\newtcolorbox{promptbox}[1]{
  enhanced,
  breakable,
  colback=black!2,
  colframe=black!25,
  coltitle=black,
  boxrule=0.4pt,
  arc=0.8mm,
  left=2mm,
  right=2mm,
  top=1.5mm,
  bottom=1.5mm,
  title={#1},
  fonttitle=\bfseries\small,
  before skip=6pt,
  after skip=8pt
}

\begin{document}

\title{NutriVisRAG: Multimodal Exemplar Grounding for Nutrient Estimation with Vision-Language Models}

\titlerunning{NutriVisRAG}

\author{
Lubnaa Abdur Rahman\inst{1,2}\orcidlink{0000-0001-7271-2814}
\and
Ioannis Papathanail\inst{2}\orcidlink{0000-0002-7767-3831}
\and
Maria Panagiotou\inst{1,2}\orcidlink{0009-0005-0117-033X}
\and
Nefeli Dimitropoulou\inst{3}\orcidlink{0009-0003-9555-8444}
\and
Margret Harðardóttir\inst{3}
\and
Elisabeth J. den Brok\inst{3}\orcidlink{0009-0002-5289-8070}
\and
Bastiaan E. de Galan\inst{3}\orcidlink{0000-0002-1255-7741}
\and
Stavroula Georgia Mougiakakou\inst{2}\orcidlink{0000-0002-6355-9982}
}

\authorrunning{L. Abdur Rahman et al.}

\institute{
Graduate School for Cellular and Biomedical Sciences, University of Bern, 3012 Bern, Switzerland
\and
ARTORG Center for Biomedical Engineering Research, University of Bern, 3008 Bern, Switzerland
\and
Department of Endocrinology, Maastricht University Medical Centre+, and Cardiovascular Research Institute Maastricht (CARIM), Maastricht University, Maastricht, the Netherlands
}


\maketitle

\begin{abstract}
Diet is a major determinant of health, yet accurate dietary assessment from free-living meal images remains challenging.
Existing retrieval-based Vision-Language Model (VLM) approaches to nutrient estimation typically rely on textual food representations, while, to our knowledge, direct grounding on visually similar annotated meals has not been investigated.
We present NutriVisRAG, a retrieval-augmented generation framework that uses a vision encoder to retrieve similar meals with known nutritional annotations and provide them as multimodal in-context exemplars for VLM nutrient estimation at inference time, without task-specific fine-tuning.
We evaluate NutriVisRAG across four datasets, four VLMs, three retrieval encoders, and two prompting strategies, including free-living meals from adults with insulin-treated diabetes, where carbohydrate estimates are compared against two dietitians' annotations. 
Across all 96 configurations (4 datasets $\times$ 4 VLMs $\times$ 3 retrieval encoders $\times$ 2 prompting strategies), NutriVisRAG reduced the primary estimation error in 56, with improvements more frequent under whole-meal (35/48) than per-item prompting (21/48).
Across the evaluated configurations, the lowest NutriVisRAG error was below the best Vanilla error in 13 of 16 VLM--dataset pairs.
For GPT-5.4-mini, under whole-meal prompting, the best-performing retrieval encoder for each dataset reduced the primary error on all four datasets relative to Vanilla, with reductions of 1.7--35.7\%.
Overall, our results show that grounding in multimodal exemplars comprising images and respective annotations can improve nutrient estimation at inference time without task-specific fine-tuning, although the benefits remain configuration-dependent.

\end{abstract}

\section{Introduction}

Accurate dietary assessment is central to nutritional monitoring and the prevention and management of diet-related chronic diseases, yet conventional self-reported approaches remain burdensome and prone to measurement errors \cite{dalakleidi2022applying}.
Advances in computer vision and smartphone technologies have enabled automated dietary assessment from meal images through multi-stage pipelines combining food segmentation, recognition, portion estimation, and nutrient calculation, alongside integrated models that directly predict food quantity and nutritional composition \cite{meyers2015im2calories,lu2020gofood,thames2021nutrition5k,papathanail2023gofood,qi2025vif}.
These approaches have benefited from increasingly large food segmentation and recognition datasets \cite{bossard2014food101,min2020isiafood500,min2023food2k,rodriguez2024lofi}.
Errors in either segmentation or recognition can propagate to downstream portion and nutrient estimation, and recognition remains particularly vulnerable to distributions not represented during training \cite{abdurrahman2025oodfood}.
Even when food regions and identities are correctly determined, estimating physical quantity remains a separate challenge, as portion estimation typically relies on depth, geometric reconstruction, or reference cues that require additional acquisition information \cite{meyers2015im2calories,lu2020gofood,thames2021nutrition5k,papathanail2023gofood,abdur2023comparative}.
End-to-end models instead bypass this by predicting food weight, energy, or macronutrients from visual representations.
However, these approaches require large training datasets with quantitative dietary annotations \cite{thames2021nutrition5k,qi2025vif,tanabe2025reasoning}.

Large Language Models, pretrained on large-scale text corpora, have demonstrated strong few-shot and in-context learning capabilities across diverse language tasks without task-specific fine-tuning \cite{brown2020language,bommasani2021opportunities}.
Extending this foundation-model paradigm to visual inputs, Vision-Language Models (VLMs) combine language modelling with visual representations to support multimodal understanding and generation \cite{radford2021clip,alayrac2022flamingo}.
Recent studies have evaluated VLMs for food recognition, meal decomposition, portion and weight estimation, calorie estimation, and macronutrient prediction, with performance varying substantially across models and prompting strategies \cite{abdur2026generative,romero2025vlmfood,yin2025foodlmm,ohara2025chatgpt,tanabe2025reasoning,mu2025foundation,vedovelli2026architecture,sterling2026vlmdietary}.
Task-specific fine-tuning, specialised prediction components, and parameter-efficient adaptation can improve performance, but require additional dietary training data and model adaptation \cite{abdur2026generative,yin2025foodlmm,tanabe2025reasoning}.
Nevertheless, quantitative dietary estimation remains challenging because food quantity and nutritional content must still be inferred from limited visual information \cite{ohara2025chatgpt,tanabe2025reasoning,mu2025foundation,vedovelli2026architecture,min2026comparative,bhatambarekar2026agentic,sterling2026vlmdietary}.
Explicitly supplied nutritional information can nevertheless improve nutrient estimation \cite{mu2025foundation,tanabe2025reasoning}.

Structured, retrieval-augmented, and agentic approaches have therefore been explored for VLM-based dietary assessment \cite{abdur2026generative,bhatambarekar2026agentic,coburn2026openknead}.
Retrieval-Augmented Generation (RAG) provides inference-time access to external evidence without requiring fine-tuning of VLMs \cite{lewis2020rag}.
In dietary applications, retrieval has been used to support food identification and nutrient estimation through dietary records, food composition databases, knowledge graphs, and standardised nutritional information \cite{dhaliwal2025nutribench,yan2025dietai24,zhou2026nutrirag,coburn2026openknead,abdurrahman2025swissfkg}.
However, current dietary retrieval remains predominantly textual or structured.
Image-based approaches first interpret meal images into textual descriptions, candidate foods, or decomposed food items before querying external food composition resources \cite{yan2025dietai24,coburn2026openknead}.
Multimodal RAG has shown that retrieval can operate directly over visual and heterogeneous representations \cite{chen2022murag,hu2023reveal,yu2025visrag}, while visual retrieval has previously been explored in food computing for image-recipe matching and calorie estimation \cite{salvador2017recipe1m,miyazaki2011image}.
To the best of our knowledge, however, VLM-based nutrient estimation applications have yet to investigate the retrieval of visually similar meal images with known nutritional annotations and provide them directly to the VLM for grounding.

We therefore investigate whether simple inference-time grounding with visually similar annotated meals can improve VLM-based nutrient estimation.
Our contributions are:
\begin{itemize}
\item We introduce \textbf{NutriVisRAG}, an inference-time framework that integrates image-based retrieval with pretrained VLMs, using similar annotated meals as multimodal in-context exemplars for nutrient estimation.
\item We systematically evaluate visual exemplar grounding across VLMs, retrieval encoders, prompting strategies, and four dietary datasets, including free-living meals from adults with insulin-treated diabetes evaluated against dietitians' visual estimations.
\end{itemize}

\section{Methods}

\label{sec:methods}

\subsection{NutriVisRAG}
\label{subsec:rag}
\subsubsection{Overview}
\begin{figure}[tb]
\centering
\includegraphics[width=0.9\textwidth]{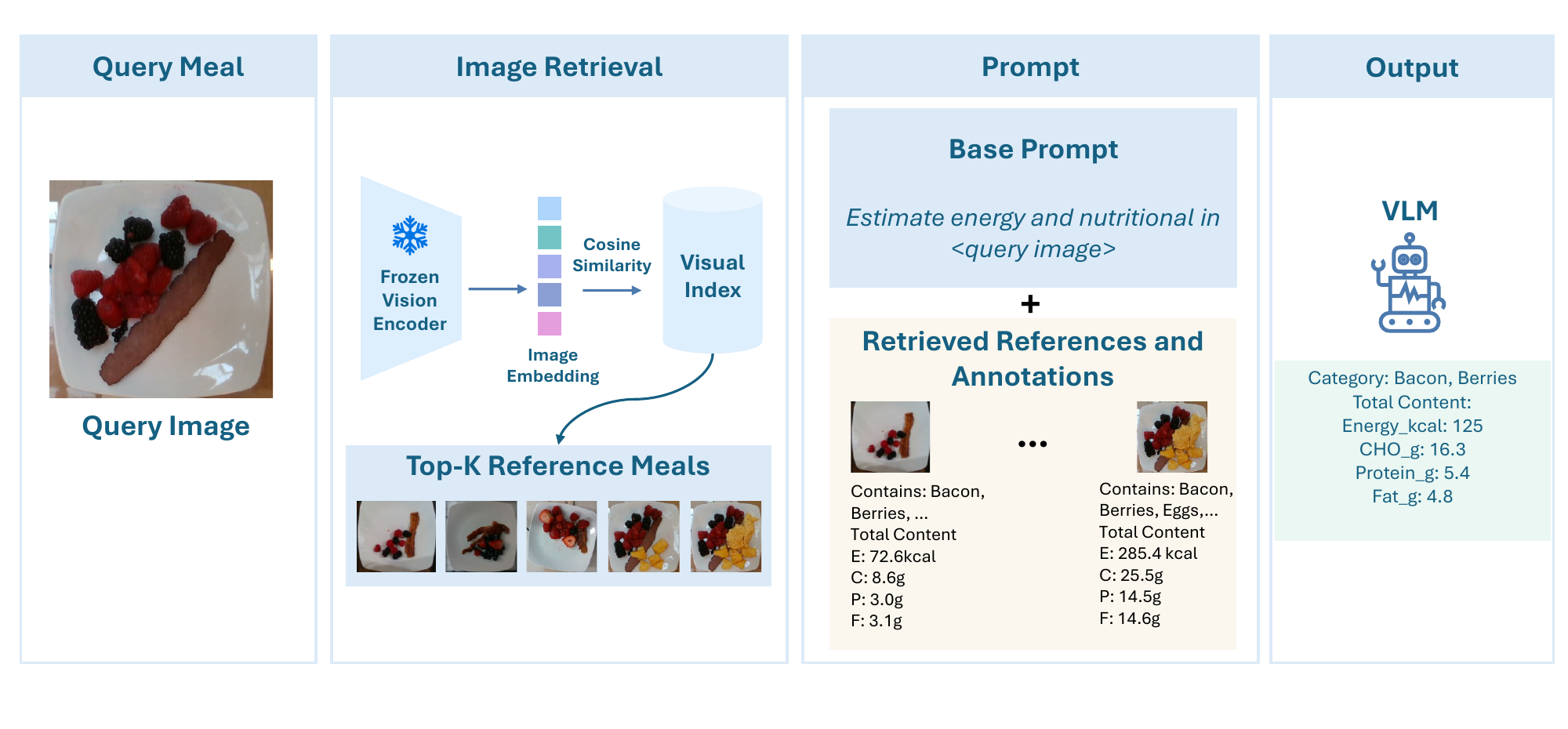}
\caption{Overview of the NutriVisRAG pipeline: the top-$K$ visually similar reference meals are retrieved for a query image, and their images and nutritional annotations are added to the prompt as multimodal context for the VLM.}
\label{fig:nutrivisrag}
\end{figure}

NutriVisRAG, illustrated in Figure~\ref{fig:nutrivisrag}, retrieves visually similar annotated meals directly in the visual embedding space and provides each retrieved meal image together with its available nutritional annotations to a VLM as multimodal context at test time.
Retrieval is followed by a single VLM generation step for nutrient estimation.
The retrieved meals therefore act as multimodal in-context exemplars carrying both visual and quantitative information \cite{lewis2020rag,alayrac2022flamingo}.

\subsubsection{Problem Formulation}
Let $I_q$ be a query meal image and $y_q$ its vector of available annotated nutritional targets, which is used only for evaluation.
Depending on the dataset, $y_q$ contains the available subset of weight, energy, carbohydrate (CHO), protein, and fat.
Let $M$ denote the VLM and $P$ the task prompt.
We compare Vanilla VLM estimation with retrieval-grounded estimation:
\begin{align}
  \text{Vanilla:} \qquad
  & \hat{y}_q = M\bigl(P,I_q\bigr),
  \label{eq:baseline}\\
  \text{NutriVisRAG:} \qquad
  & \hat{y}_q = M\bigl(P,I_q,\mathcal{E}(q)\bigr),
  \label{eq:rag}
\end{align}
where $\mathcal{E}(q)$ is a set of retrieved reference meals $(I_k,a_k)$, each consisting of a reference image $I_k$ and its available nutritional annotations $a_k$, which are provided to the VLM as input.
The two conditions therefore differ only by the retrieved exemplar context $\mathcal{E}(q)$, whose construction is described below.

\paragraph{Index construction.}
Let $\phi(\cdot)$ denote the frozen visual encoder used for retrieval.
Each reference image $I_j$ is mapped to an embedding $\phi(I_j)$ and stored in the visual index together with its available nutritional annotations.
Evaluation query images are excluded from the reference index.

\paragraph{Retrieval.}
At test time, the query image is embedded using the same visual encoder, and its similarity to each indexed reference image $I_j$ is computed using cosine similarity:
\begin{equation}
  s(q,j)
  =
  \frac{\phi(I_q)^{\top}\phi(I_j)}
  {\lVert \phi(I_q)\rVert\,\lVert \phi(I_j)\rVert}.
  \label{eq:similarity}
\end{equation}
The $K$ references with the highest similarity form the candidate set $\mathcal{E}_K(q)=\{(I_k,a_k)\}_{k=1}^{K}$, ordered by decreasing similarity, where $s_k$ denotes the similarity of the $k$-th ranked reference.
A minimum similarity threshold $\tau$ is applied, and if no candidate satisfies it, the highest-ranked reference is retained as a fallback:
\begin{equation}
  \mathcal{E}(q) =
  \begin{cases}
    \bigl\{(I_k,a_k) \in \mathcal{E}_K(q) : s_k \ge \tau\bigr\}, & \text{if this set is non-empty},\\[2pt]
    \bigl\{(I_1,a_1)\bigr\}, & \text{otherwise}.
  \end{cases}
  \label{eq:final-set}
\end{equation}

Thresholding may therefore retain fewer than $K$ exemplars, while the fallback ensures that the VLM always receives at least the nearest reference for grounding.
The annotations accompanying each retrieved image are provided at item and meal level where available; unavailable quantities are omitted rather than imputed (e.g., SNAPMe references are presented without meal weight).

\subsection{Datasets}
\label{subsec:datasets}

We use four dietary image datasets (Table~\ref{tab:datasets}).

\textit{Nutrition5k}~\cite{thames2021nutrition5k} contains cafeteria meals acquired under controlled conditions, with ingredient-level food identities and weights and meal-level weight, energy, and macronutrient annotations.

\textit{SNAPMe}~\cite{larke2023snapme} contains free-living meal images captured from smartphones and linked to dietary records collected using the Automated Self-Administered 24-hour Dietary Assessment Tool (ASA24), providing verified self-reported annotations for energy and macronutrients without measured meal weight.

\textit{MADiMa}~\cite{allegra2017multimedia} contains 80 meals with two to four food items and measured weight, volume, and energy and macronutrient composition.
Each meal was captured at 40\,cm and 60\,cm under a constrained acquisition protocol, yielding 160 images used jointly in the main evaluation. 
The two acquisition distances are additionally analysed as separate 80-image subsets in Supplementary Section~S2.

\textit{EDUCATE}~\cite{denbrok2025educate} contains free-living meal images captured using the goFOOD\textsuperscript{\texttrademark} Lite mobile application ~\cite{lu2020gofood} from adults with insulin-treated diabetes in the Netherlands arm of the study \cite{denbrok2025educate}.
After excluding packaged foods, erroneous entries, and non-food images, de-identified food-only images from 25 participants were retained.
CHO content was independently estimated by two dietitians using an annotation tool~\cite{abdurrahman2024mealsam}, and model estimates are evaluated against their mean in the absence of measured ground truth.

Nutrition5k and SNAPMe serve as both evaluation query sets and indexed reference corpora.
For each of Nutrition5k and SNAPMe, a subset (exact numbers in Table~\ref{tab:datasets}) is held out for evaluation, and the remaining are combined into the reference corpus, ensuring that evaluated meals cannot be retrieved as references.
Despite annotation provenance differences, the two reference subsets are combined to broaden the range of meal compositions and acquisition conditions represented, spanning controlled and free-living settings.
Nutrition5k queries are drawn from overhead, non-empty meals, while SNAPMe queries are selected round-robin at image level with preference for compositionally rich plates.
The task-specific Nutrition5k split does not follow the official partition and was designed to retain compositionally richer dishes in the reference corpus.

\begin{table}[tb]
\caption{Datasets used.}
\label{tab:datasets}
\centering
\footnotesize
\resizebox{0.9\textwidth}{!}{%
\begin{tabular}{lccc}
\toprule
\textbf{Dataset} & \textbf{Evaluation queries} & \textbf{Indexed references} & \textbf{Evaluation targets} \\
\midrule
Nutrition5k~\cite{thames2021nutrition5k} & 815 & 2,445 & Weight, energy, CHO, protein, fat \\
SNAPMe~\cite{larke2023snapme}            & 323 & 1,149  & Energy, CHO, protein, fat \\
MADiMa~\cite{allegra2017multimedia}      & 160 (80 meals each at 40 and 60\,cm) & -- & Weight, energy, CHO, protein, fat \\
EDUCATE~\cite{denbrok2025educate}        & 247 & -- & CHO \\
\bottomrule
\end{tabular}}
\end{table}

\subsection{Experimental Setup}
\label{subsec:impl}

\paragraph{Models and retrieval encoders.}
We evaluate the four VLMs listed in Table~\ref{tab:vlms} without task-specific fine-tuning.
The open-weight local models are run on an NVIDIA RTX A6000 through Ollama \cite{ollama}, the hosted open-weight models through Ollama Cloud, and GPT through the OpenAI API \cite{openai_api}.
Retrieval uses the three frozen image encoders listed in Table~\ref{tab:encoders}.
\begin{table}[tb]
\centering

\begin{minipage}[t]{0.49\textwidth}
\centering
\scriptsize
\caption{Vision--language models evaluated.}
\label{tab:vlms}
\resizebox{\linewidth}{!}{%
\begin{tabular}{llll}
\toprule
\textbf{Model} & \textbf{Identifier} & \textbf{Weights} & \textbf{Serving} \\
\midrule
GPT-5.4-mini~\cite{gpt}
& \texttt{gpt-5.4-mini}
& proprietary
& OpenAI API \\

Gemma-4-31B~\cite{gemma}
& \texttt{gemma4:31b-cloud}
& open
& Ollama Cloud \\

Mistral-Small-3.2~\cite{mistral}
& \texttt{mistral-small3.2:24b}
& open
& local Ollama \\

GLM-5.3-Flash~\cite{zai_glm53flash}
& \texttt{glm-5.3-flash:cloud}
& open
& Ollama Cloud \\
\bottomrule
\end{tabular}%
}
\end{minipage}
\hfill
\begin{minipage}[t]{0.49\textwidth}
\centering
\scriptsize
\caption{Frozen image encoders used for retrieval.}
\label{tab:encoders}
\resizebox{\linewidth}{!}{%
\begin{tabular}{lll}
\toprule
\textbf{Encoder} & \textbf{Checkpoint} & \textbf{Input resolution} \\
\midrule
PE-Core-G14~\cite{bolya2025pe}
& \texttt{facebook/PE-Core-G14-448}
& $448{\times}448$ \\

SigLIP\,2~\cite{tschannen2025siglip2}
& \texttt{google/siglip2-so400m-patch14-384}
& $384{\times}384$ \\

Nomic Vision~\cite{nussbaum2024nomic}
& \texttt{nomic-ai/nomic-embed-vision-v1.5}
& $224{\times}224$ \\
\bottomrule
\end{tabular}%
}
\end{minipage}

\end{table}

\paragraph{Prompting strategies.}
We evaluate two prompting strategies differing in the granularity of the requested output.
In \textit{per-item}, the VLM was asked to identify the visible food items and estimate the weight, energy, CHO, protein, and fat of each item separately, with meal-level estimates obtained by summing the returned item-level predictions.
In \textit{whole-meal}, the VLM was asked to directly estimate the corresponding totals for the complete meal without explicit item-level decomposition.
For each strategy, the same task instructions and enforced structured JSON output format are used in the Vanilla and grounded conditions.
The grounded condition additionally receives the retrieved reference images with their nutritional annotations and a system-prompt description of this reference information.
The exact prompt templates are provided in Supplementary Section~S1.

\paragraph{Retrieval configuration.}
The Nutrition5k and SNAPMe reference sets are combined into a single retrieval corpus, with evaluation query images excluded from the index.
Visual indices are built using Facebook AI Similarity Search (FAISS) through LlamaIndex~\cite{johnson2019faiss,liu2022llamaindex}.
Reference images are embedded once for each encoder and $\ell_2$-normalised before retrieval.
The main experiments use $K=5$, a minimum cosine-similarity threshold of $\tau=0.75$ with the fallback strategy defined above, and no reranking.
The values $K=5$ and $\tau=0.75$ were selected from preliminary comparisons with $K=10$ and $\tau=0$ on fixed 100-query subsets of each query set.

\paragraph{Retrieval ablations.}
We report the fixed 100-query comparisons used to examine retrieval-parameter sensitivity and select the main retrieval settings.
The same queries are reused across configurations to enable paired comparisons under identical evaluation samples.
We evaluate these comparisons across all four VLMs under both per-item and whole-meal prompting.
We vary the similarity threshold and retrieval depth relative to the default retrieval configuration:
\begin{itemize}
\item \textbf{Similarity threshold:} $\tau=0$ versus $\tau=0.75$.
\item \textbf{Number of retrieved exemplars:} $K=5$ versus $K=10$.
\end{itemize}

\paragraph{Exploratory retrieval extensions.}
Beyond these retrieval-parameter ablations, we evaluate two exploratory extensions on the same fixed 100-query subsets.
Both extensions are evaluated across all four VLMs.
For post-hoc food-composition grounding, we match VLM-predicted food names to food-composition database entries using lexical token-set similarity or one of three text-embedding models: Qwen3-Embedding~\cite{zhang2025qwen3}, EmbeddingGemma~\cite{vera2025embeddinggemma}, and Nomic Embed Text v2 (MoE)~\cite{nussbaum2025training}.
The matched nutrient values are scaled by the VLM-predicted food weights and summed to obtain meal-level estimates.
This analysis uses per-item predictions, which provide the food identities and corresponding weights required for database lookup.
For retrieval reranking, each query retrieves 50 visual candidates, which are rescored using Jina Reranker M0~\cite{jina2025rerankerm0}, Qwen3-VL-Reranker-2B~\cite{li2026qwen3vlreranker}, or Llama Nemotron Rerank VL 1B v2~\cite{nvidia2025nemotronrerankvl}.
The five highest-ranked references are then provided to the VLM under both prompting strategies.
These analyses test whether more selective exemplar ranking or external food-composition lookup improves the base NutriVisRAG estimates, respectively.
Full experimental details and results are reported in Supplementary Section~S5.

\paragraph{Inference.}
Query and retrieved reference images are resized to a maximum side length of 800 pixels while preserving aspect ratio, providing a common input scale without upscaling the lowest-resolution images and limiting visual input size.
Temperature $0$ and a fixed seed of $42$ were used to minimise sampling variability, although deterministic outputs are not guaranteed across serving backends.
Repeated inference runs were not performed, so run-to-run variance was not independently quantified.

\subsection{Evaluation}
\label{subsec:eval}

For each nutritional field $v \in \{\mathrm{weight},\mathrm{energy},\mathrm{CHO},\mathrm{protein},\mathrm{fat}\}$, let $y_{i,v}$ and $\hat{y}_{i,v}$ denote the target and predicted values for meal $i = 1,\dots,N$.
We report the mean absolute error normalised by the mean target value,
\begin{equation}
  \mathrm{\%MAE}_v
  =
  \frac{\frac{1}{N}\sum_{i=1}^{N}\left|\hat{y}_{i,v}-y_{i,v}\right|}
       {\frac{1}{N}\sum_{i=1}^{N}y_{i,v}}
  \times 100.
  \label{eq:pmae}
\end{equation}
Normalising by the mean rather than per meal avoids unstable percentage errors when individual target values are near zero, while keeping nutritional fields of different magnitudes on a comparable scale.

Macronutrient performance is summarised in two ways:
$\mathrm{\%MAE}_{\text{macro}}$ is the unweighted mean of the three macronutrient field errors, giving each equal weight regardless of magnitude, while $\mathrm{\%MAE}_{\text{micro}}$ pools absolute errors in grams across the three fields before normalising, and is therefore dominated by the macronutrient present in the largest quantity,

\begin{equation}
\resizebox{0.9\linewidth}{!}{$\displaystyle
\mathrm{\%MAE}_{\text{macro}} =
\tfrac{1}{3}\!\sum_{v\in\{c,p,f\}}\mathrm{\%MAE}_v,
\qquad
\mathrm{\%MAE}_{\text{micro}} =
\frac{\sum_{v\in\{c,p,f\}}\sum_i
|\hat{y}_{i,v}-y_{i,v}|}
{\sum_{v\in\{c,p,f\}}\sum_i y_{i,v}}
\times 100,
$}
\end{equation}
where $c$, $p$, and $f$ denote CHO, protein, and fat, respectively.
On EDUCATE, where only CHO is annotated, $\mathrm{\%MAE}_{\text{CHO}}$ is computed against the dietitians' mean annotations.
We use $\mathrm{\%MAE}_{\text{micro}}$ as the primary metric, replaced by $\mathrm{\%MAE}_{\text{CHO}}$ on EDUCATE, and denote this as $\mathrm{\%MAE}_{\text{micro/CHO}}$.

\paragraph{Statistical testing.}
Methods are compared using two-sided paired Wilcoxon signed-rank tests on per-meal absolute errors \cite{woolson2007wilcoxon}.
Multiple comparisons are corrected using the Benjamini--Hochberg procedure within each experiment--outcome family, controlling the false discovery rate at $q=0.05$ \cite{chen2017general}.
A family comprises all comparisons of one outcome measure within one experiment, pooled across query sets, VLMs, encoders, and prompting strategies; for example, all $\mathrm{\%MAE}_{\text{micro/CHO}}$ comparisons in the threshold ablation form one family, and all $\mathrm{\%MAE}_{\text{micro/CHO}}$ comparisons of NutriVisRAG against Vanilla form another.

\paragraph{Retrieval diagnostics.}
Using the full query sets, retrieval benefit is the per-query reduction in pooled absolute macronutrient error from Vanilla to NutriVisRAG, normalised by the query-set mean target macronutrient sum; EDUCATE instead uses absolute CHO error normalised by mean annotated CHO.
Visual similarity is the mean cosine similarity of the retrieved exemplars, while nutritional distance is computed retrospectively as the mean absolute query--exemplar difference across the available macronutrients, each normalised by its query-set mean.
Associations are quantified by query set, retrieval encoder, and prompting strategy using Spearman's rank correlation, with retrieval benefit averaged across the four VLMs; per-VLM results are in Supplementary Section~S4.2.

\section{Results and Discussion}

\subsection{NutriVisRAG versus Vanilla VLM Inference}
\label{subsec:main-results}

We first evaluate whether visually retrieved meals with known nutritional annotations improve nutrient estimation relative to Vanilla VLM inference.
Table~\ref{tab:rag-results} reports the primary error, with complete per-nutrient results in Supplementary Table~S1.
Across 96 configurations (4 datasets $\times$ 4 VLMs $\times$ 3 retrieval encoders $\times$ 2 prompting strategies), NutriVisRAG reduces the corresponding Vanilla error in 56, with improvements more frequent under whole-meal (35/48) than per-item prompting (21/48).
After Benjamini--Hochberg correction, 45 configurations show a significant reduction in error, while 23 show a significant increase.
Whole-meal prompting may benefit more from retrieval since exemplar annotations directly match the requested meal-level totals, unlike per-item prompting, which requires additional decomposition and aggregation.

Using the primary $\mathrm{\%MAE}_{\text{micro/CHO}}$ metric, NutriVisRAG reduces error in 13/24 configurations on Nutrition5k, 7/24 on SNAPMe, 22/24 on MADiMa, and 14/24 on EDUCATE.
Nutrient-specific effects are broader, particularly for CHO, with improvements in 18/24, 9/24, 24/24, and 14/24 configurations, respectively; on MADiMa, weight and energy also improve in 21/24 and 19/24 configurations (Supplementary Table~S1).
Despite contributing indexed references, SNAPMe benefits least consistently, while MADiMa and EDUCATE show more frequent gains despite stronger projected separation from the reference corpus under some encoders (Supplementary Fig.~S2).
Thus, corpus membership and projected embedding overlap alone do not determine retrieval effectiveness.
SNAPMe and EDUCATE should be interpreted with their annotation sources in mind: SNAPMe uses verified self-reported dietary estimates, whereas EDUCATE is restricted to CHO compared to dietitians' mean estimates, which are not nutritional ground truth.
Across the evaluated configurations, the lowest NutriVisRAG error is below the best Vanilla error in 13 of 16 VLM--query-set pairs.
As a retrieval-only control, we predict each nutritional target as the cosine-similarity-weighted mean of the retained exemplar annotations, without VLM inference.
The lowest NutriVisRAG error is below the best retrieval-only error on all four query sets, by 5.8, 8.3, 12.7, and 7.1 percentage points on Nutrition5k, SNAPMe, MADiMa, and EDUCATE, respectively.
Thus, direct similarity-weighted neighbour averaging does not account for the lowest observed NutriVisRAG errors.

The response to retrieval is strongly VLM-dependent, with improvements in 19/24 configurations for GPT-5.4-mini, 16/24 for GLM-5.3-Flash, 14/24 for Gemma-4-31B, and 7/24 for Mistral-Small-3.2.
Mistral-Small-3.2 shows substantial error increases under per-item prompting on Nutrition5k and SNAPMe but benefits from whole-meal grounding on Nutrition5k and MADiMa, whereas GPT-5.4-mini benefits more consistently across prompting strategies on Nutrition5k, MADiMa, and EDUCATE and GLM-5.3-Flash shows strong gains on MADiMa and EDUCATE.
Model scale alone does not explain these differences: Mistral-Small-3.2 is a 24B dense model, Gemma-4-31B has 31B parameters, GLM-5.3-Flash has 320B total and 18B active parameters, and GPT-5.4-mini's parameter count is undisclosed, while architecture, training data, and post-training are not independently controlled~\cite{mistral,gemma,zai_glm53flash,gpt}.

No retrieval encoder consistently dominates across individual model--query-set--prompt configurations (Table~\ref{tab:rag-results} and Supplementary Table~S1).
Although cross-dataset means (Supplementary Table~S2) favour PE-Core in several VLM--prompt combinations, the best encoder changes across query sets, and the direction of the retrieval effect can also change for the same VLM and prompting strategy.
Overall, NutriVisRAG provides clear but configuration-dependent gains, while improvements in $\mathrm{\%MAE}_{\text{micro/CHO}}$ do not extend to every nutritional target.
Median query latency increases by $1.4\times$ for GPT-5.4-mini, $1.7\times$ for Gemma-4-31B, $1.3\times$ for Mistral-Small-3.2, and $1.2\times$ for GLM-5.3-Flash, corresponding to $+0.7$, $+1.2$, $+3.3$, and $+1.0$\,s per meal, respectively.

\begin{figure}[tb]
\centering
\includegraphics[width=0.5\textwidth]{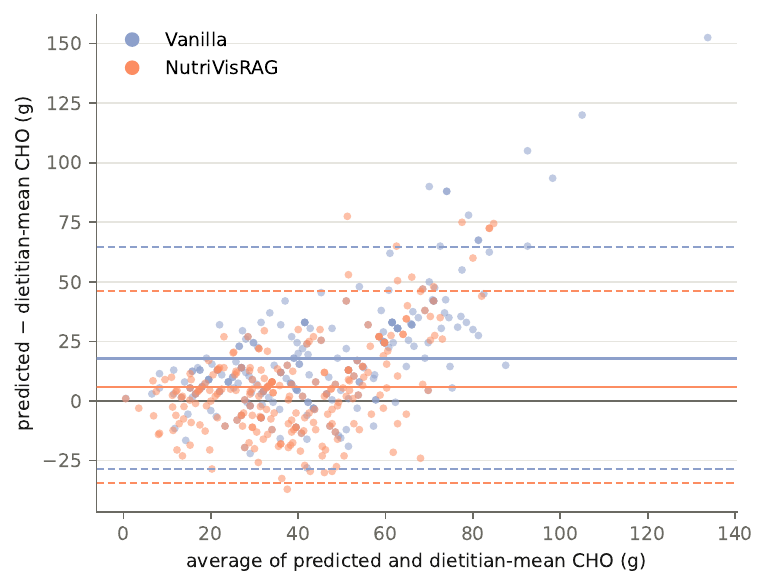}
\caption{Bland--Altman analysis of CHO estimates for 247 EDUCATE meals using GPT-5.4-mini with SigLIP\,2 under whole-meal prompting, without (blue) and with (orange) NutriVisRAG.
Differences are computed against the mean dietitian estimate; solid and dashed lines show the mean bias and 95\% limits of agreement, respectively.}
\label{fig:educate-ba}
\end{figure}

Given the mixed results on EDUCATE (14/24 improvements) and its relevance to free-living CHO estimation in people with diabetes, we further examine whether retrieval improves meal-level agreement with dietitian estimates beyond error.
Figure~\ref{fig:educate-ba} presents GPT-5.4-mini with SigLIP\,2 under whole-meal prompting, for which NutriVisRAG significantly reduces $\mathrm{\%MAE}_{\text{CHO}}$ from 58.4 to 43.3.
NutriVisRAG reduces mean overestimation from 18.0 to 5.9\,g and narrows the 95\% limits of agreement from $[-28.5,64.5]$ to $[-34.3,46.1]$\,g, particularly mitigating large positive errors at higher CHO values.
Although individual errors remain, an improved agreement with dietitian estimates is seen; complete analyses across all query sets, VLMs, and encoders are provided in Supplementary Section~S4.4.

\begin{table}[tb]
\centering
\caption{Vanilla versus NutriVisRAG for all retrieval encoders and both prompting strategies, on the full query sets.
Values are $\mathrm{\%MAE}_{\text{micro}}$ ($\mathrm{\%MAE}_{\text{CHO}}$ for EDUCATE); lower is better.
Bold: lowest error within each prompting strategy per model and query set.
$^\star$/$^{\blacktriangledown}$: significant improvement/degradation over the corresponding Vanilla baseline (paired Wilcoxon signed-rank test, Benjamini--Hochberg corrected, $q<0.05$).}
\label{tab:rag-results}
\scriptsize
\setlength{\tabcolsep}{3pt}

\resizebox{0.9\textwidth}{!}{%
\begin{tabular}{llrrrrrrrr}
\toprule
& & \multicolumn{4}{c}{\textbf{Per-item}} & \multicolumn{4}{c}{\textbf{Whole-meal}} \\
\cmidrule(lr){3-6}\cmidrule(lr){7-10}
& & & \multicolumn{3}{c}{\textbf{NutriVisRAG}} & & \multicolumn{3}{c}{\textbf{NutriVisRAG}} \\
\cmidrule(lr){4-6}\cmidrule(lr){8-10}
\textbf{Query set} & \textbf{Model} & \textbf{Vanilla} & \textbf{PE-Core} & \textbf{SigLIP2} & \textbf{Nomic Vision} & \textbf{Vanilla} & \textbf{PE-Core} & \textbf{SigLIP2} & \textbf{Nomic Vision} \\
\midrule
\multirow{4}{*}{Nutrition5k} & Gemma-4-31B & \textbf{53.6} & 53.7 & 54.6 & 59.8 & 63.5 & 51.0$^\star$ & \textbf{50.8}$^\star$ & 55.8$^\star$ \\
& GPT-5.4-mini & 58.0 & \textbf{42.7}$^\star$ & 43.4$^\star$ & 47.1$^\star$ & 58.1 & \textbf{43.4}$^\star$ & 44.2$^\star$ & 48.0$^\star$ \\
& Mistral-Small-3.2 & \textbf{72.1} & 111.2$^{\blacktriangledown}$ & 119.5$^{\blacktriangledown}$ & 145.1$^{\blacktriangledown}$ & 78.7 & \textbf{62.3}$^\star$ & 63.3$^\star$ & 74.3$^\star$ \\
& GLM-5.3-Flash & 54.0 & 54.1$^{\blacktriangledown}$ & \textbf{53.4}$^\star$ & 58.6$^{\blacktriangledown}$ & \textbf{56.2} & 58.8$^{\blacktriangledown}$ & 56.8$^{\blacktriangledown}$ & 62.2$^{\blacktriangledown}$ \\
\midrule
\multirow{4}{*}{SNAPMe} & Gemma-4-31B & \textbf{41.0} & 46.2$^{\blacktriangledown}$ & 44.0 & 45.0$^{\blacktriangledown}$ & 43.6 & 42.5 & \textbf{40.4}$^\star$ & 41.9 \\
& GPT-5.4-mini & \textbf{41.4} & 41.7 & 42.2 & 42.8 & 41.7 & 42.8 & \textbf{41.0} & 43.0 \\
& Mistral-Small-3.2 & \textbf{51.1} & 103.6$^{\blacktriangledown}$ & 130.3$^{\blacktriangledown}$ & 108.8$^{\blacktriangledown}$ & \textbf{44.6} & 67.2$^{\blacktriangledown}$ & 75.0$^{\blacktriangledown}$ & 66.0$^{\blacktriangledown}$ \\
& GLM-5.3-Flash & \textbf{39.8} & 40.1 & 43.8 & 41.5 & 39.6 & \textbf{38.0}$^\star$ & 38.5 & 38.8 \\
\midrule
\multirow{4}{*}{MADiMa} & Gemma-4-31B & 54.7 & 50.5$^\star$ & \textbf{47.5}$^\star$ & 50.9$^\star$ & 70.7 & 62.7$^\star$ & 63.5$^\star$ & \textbf{60.3}$^\star$ \\
& GPT-5.4-mini & 57.4 & \textbf{44.5}$^\star$ & 46.5$^\star$ & 46.4$^\star$ & 75.7 & \textbf{48.7}$^\star$ & 52.9$^\star$ & 51.6$^\star$ \\
& Mistral-Small-3.2 & 79.5 & \textbf{67.6}$^\star$ & 78.6 & 82.8 & 63.4 & \textbf{52.8}$^\star$ & 55.6$^\star$ & 69.2 \\
& GLM-5.3-Flash & 51.5 & 40.9$^\star$ & \textbf{39.7}$^\star$ & 42.8$^\star$ & 48.2 & 40.2$^\star$ & \textbf{38.9}$^\star$ & 42.2$^\star$ \\
\midrule
\multirow{4}{*}{EDUCATE} & Gemma-4-31B & \textbf{33.0} & 36.5$^{\blacktriangledown}$ & 37.8 & 38.5 & 45.4 & 45.0 & \textbf{44.5} & 47.0 \\
& GPT-5.4-mini & 45.0 & 40.9 & \textbf{38.0}$^\star$ & 42.5 & 58.4 & 46.6$^\star$ & \textbf{43.3}$^\star$ & 43.6$^\star$ \\
& Mistral-Small-3.2 & \textbf{51.2} & 153.8$^{\blacktriangledown}$ & 151.6$^{\blacktriangledown}$ & 147.1$^{\blacktriangledown}$ & \textbf{51.9} & 82.2$^{\blacktriangledown}$ & 92.8$^{\blacktriangledown}$ & 81.4$^{\blacktriangledown}$ \\
& GLM-5.3-Flash & 53.6 & \textbf{47.8}$^\star$ & 47.9$^\star$ & 53.2 & 54.0 & 44.0$^\star$ & 44.1$^\star$ & \textbf{42.0}$^\star$ \\
\bottomrule
\end{tabular}%
}
\end{table}

\subsection{Effect of Prompting Strategy}
\label{subsec:prompting}

\begin{figure}[tb]
\centering
\includegraphics[width=0.8\textwidth]{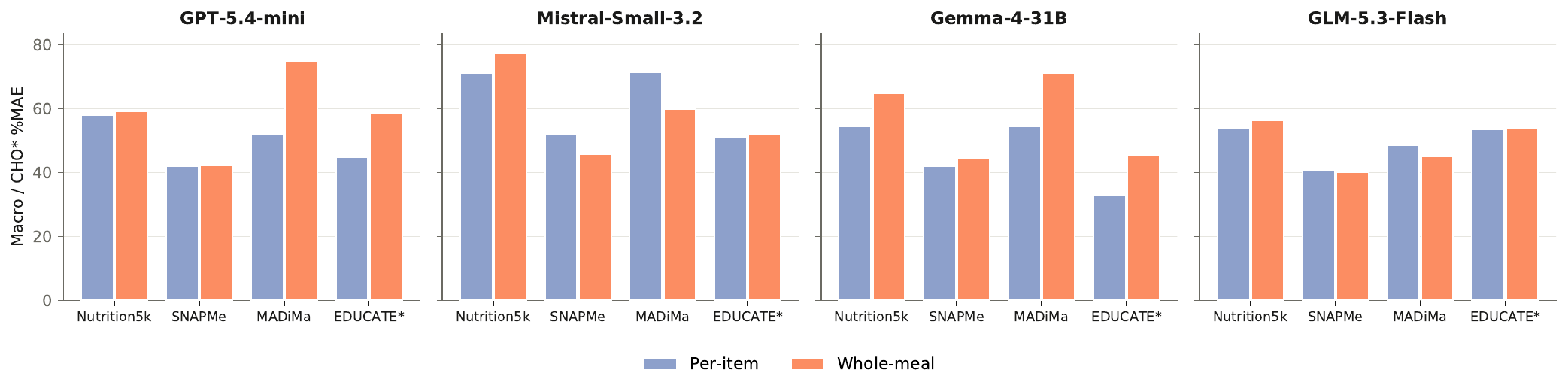}
\caption{Effect of prompting granularity under Vanilla VLM inference.
$\mathrm{\%MAE}_{\text{macro}}$ (lower is better) is reported across VLMs and query sets; (*) for EDUCATE, $\mathrm{\%MAE}_{\text{CHO}}$ against the mean dietitian estimate is shown.}
\label{fig:macrobars}
\end{figure}

Figure~\ref{fig:macrobars} compares per-item and whole-meal prompting under Vanilla VLM inference.
Per-item yields lower $\mathrm{\%MAE}_{\text{macro}}$ ($\mathrm{\%MAE}_{\text{CHO}}$ for EDUCATE) in 12/16 model--dataset pairs, while whole-meal yields lower error for Mistral-Small-3.2 and GLM-5.3-Flash on SNAPMe and MADiMa.
On MADiMa, 60\,cm images yield lower $\mathrm{\%MAE}_{\text{macro}}$ and $\mathrm{\%MAE}_{\text{micro}}$ for GPT-5.4-mini, Gemma-4-31B, and GLM-5.3-Flash under both prompting strategies and for Mistral-Small-3.2 under whole-meal prompting, but not under per-item prompting (Supplementary Fig.~S1).
Although per-item prompting performs better under Vanilla inference in most pairs, NutriVisRAG improves more frequently under whole-meal prompting, showing that stronger baseline prompt performance does not necessarily imply greater retrieval benefit.

\subsection{Exemplar Similarity and Retrieval Benefit}
\label{subsec:retrieval-benefit}

\begin{figure}[tb]
\centering
\includegraphics[width=0.7\textwidth]{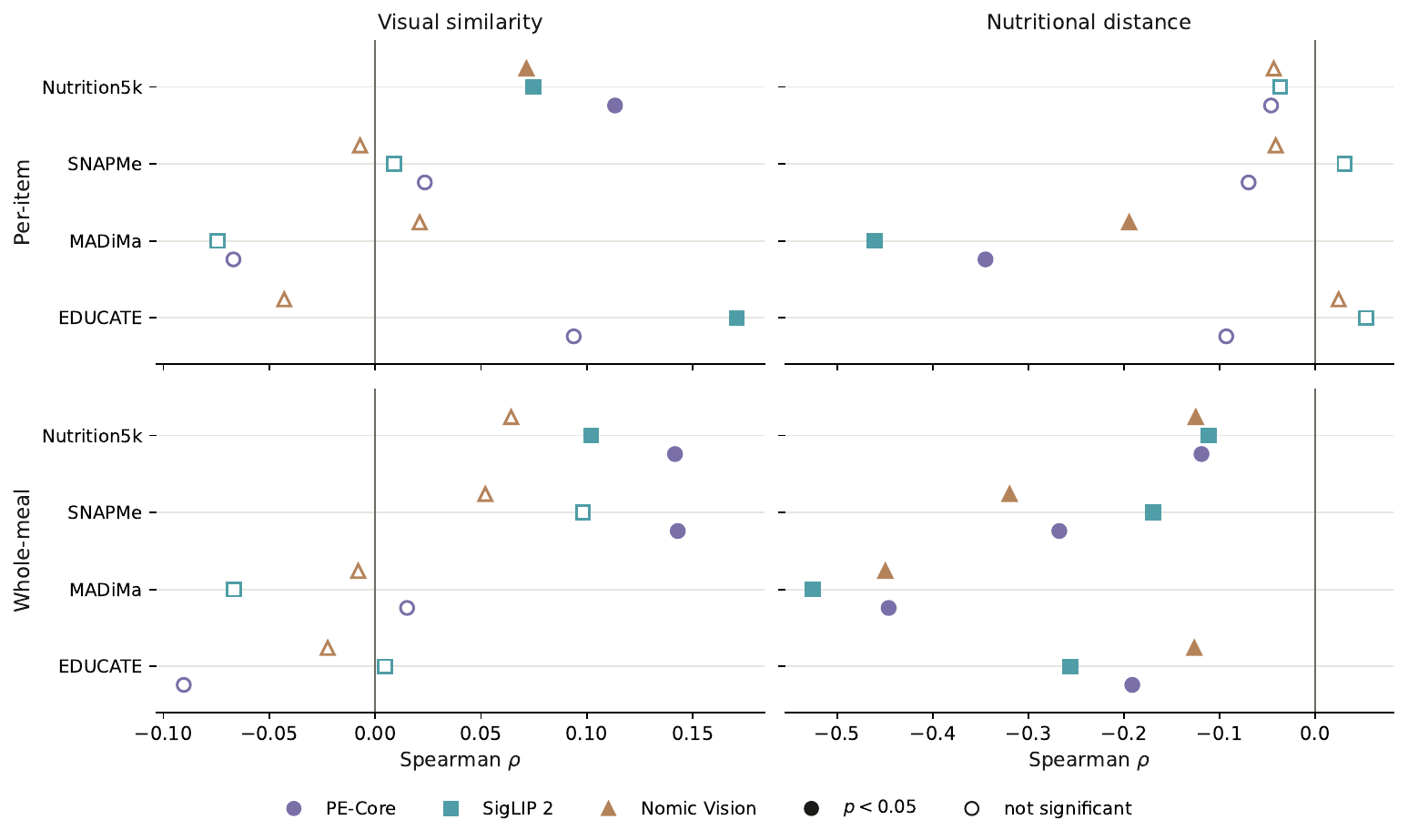}
\caption{Association of exemplar similarity with NutriVisRAG retrieval benefit under per-item and whole-meal prompting.
Spearman correlations ($\rho$) relate retrieval benefit to mean visual similarity (left) and retrospective nutritional distance (right), by query set and retrieval encoder.
Filled markers indicate Benjamini--Hochberg-corrected $q<0.05$; open markers indicate non-significant associations.
Positive visual-similarity correlations indicate greater benefit for more visually similar exemplars, whereas negative nutritional-distance correlations indicate greater benefit for nutritionally closer exemplars.}
\label{fig:retrieval-benefit}
\end{figure}

Since retrieval improves estimation in many but not all configurations, we examine whether its benefit is associated with exemplar similarity.
A preliminary t-SNE analysis (Supplementary Fig.~S2) showed that greater projected query--reference overlap does not correspond to more consistent gains, with MADiMa benefiting strongly despite its separation under PE-Core and SigLIP\,2.
At the individual-query level, Figure~\ref{fig:retrieval-benefit} relates retrieval benefit to visual similarity and retrospective nutritional distance under both prompting strategies.
Visual similarity shows weak and inconsistent associations with retrieval gains across query sets and encoders.
In contrast, nutritional distance is more consistently negatively associated with benefit under whole-meal prompting, most strongly on MADiMa, and less so under per-item prompting.
Visual similarity also corresponds to nutritional proximity more clearly on Nutrition5k than on the other query sets (Supplementary Fig.~S5).
Together, projected embedding overlap and cosine similarity alone therefore do not reliably predict exemplar usefulness.
The nutritional-distance analysis is retrospective, as query annotations are unavailable at inference time and are shared with the error metric, so the association is descriptive rather than causal.

\subsection{Retrieval Ablations}

\subsubsection{Effect of Similarity Threshold}
Table~\ref{tab:retrieval-sensitivity} compares the default similarity threshold ($\tau=0.75$) with no threshold ($\tau=0$) under both prompting strategies.
Compared with $\tau=0$, the default yields lower $\mathrm{\%MAE}_{\text{micro/CHO}}$ in 60/96 configurations, including 32/48 per-item and 28/48 whole-meal configurations.
Lower error is seen in 10/24 configurations on Nutrition5k, 18/24 on SNAPMe, 13/24 on MADiMa, and 19/24 on EDUCATE, with seven significant comparisons all favouring the default threshold.
Where thresholding does not alter the retained exemplar set, observed differences cannot be attributed to exemplar filtering and may instead reflect inference variability.
Similarity filtering benefit many configurations, but its effect remains model-, dataset-, and encoder-dependent, and cosine similarity alone does not guarantee nutritional relevance.

\subsubsection{Effect of Retrieval Depth}
\label{subsec:k-ablation}

Compared with $K=10$, the default depth ($K=5$) reduces $\mathrm{\%MAE}_{\text{micro/CHO}}$ in 56/96 configurations, including 18/24 on Nutrition5k, 14/24 on SNAPMe, 10/24 on MADiMa, and 14/24 on EDUCATE.
Of the five significant comparisons, four favour $K=5$ and all occur under per-item prompting, whereas the single comparison favouring $K=10$ occurs for Mistral-Small-3.2 with PE-Core on EDUCATE under whole-meal prompting.
Increasing retrieval depth also raises latency for every VLM, by $0.1$--$2.0$\,s per query under whole-meal and $0.4$--$18.6$\,s under per-item prompting.
Overall, increasing retrieval depth provides no consistent accuracy benefit while incurring additional inference cost, supporting $K=5$ as the default.

\begin{table*}[tb]
\centering
\caption{Retrieval ablations on the fixed 100-query subsets, with per-item and whole-meal prompting side by side (\%MAE, lower is better).
Default is the main configuration ($\tau=0.75$, $K=5$); the other columns remove the threshold ($\tau=0$) or increase depth ($K=10$).
$\mathrm{\%MAE}_{\text{micro/CHO}}$ is $\mathrm{\%MAE}_{\text{micro}}$ for Nutrition5k, SNAPMe, and MADiMa and $\mathrm{\%MAE}_{\text{CHO}}$ for EDUCATE; dashes mark unavailable targets.
Bold: lowest error within each group of three.
$^\star$/$^{\blacktriangledown}$: significantly lower/higher error than Default (paired Wilcoxon signed-rank test, Benjamini--Hochberg corrected, $q<0.05$).}
\label{tab:retrieval-sensitivity}
\scriptsize
\setlength{\tabcolsep}{1.6pt}
\renewcommand{\arraystretch}{0.95}
\resizebox{0.95\textwidth}{!}{%
\begin{tabular}{lll rrr rrr rrr rrr rrr rrr}
\toprule
& & & \multicolumn{9}{c}{\textbf{Per-item}} & \multicolumn{9}{c}{\textbf{Whole-meal}} \\
\cmidrule(lr){4-12}\cmidrule(lr){13-21}
& & & \multicolumn{3}{c}{$\mathrm{\%MAE}_{\text{weight}}$} & \multicolumn{3}{c}{$\mathrm{\%MAE}_{\text{energy}}$} & \multicolumn{3}{c}{$\mathrm{\%MAE}_{\text{micro/CHO}}$} & \multicolumn{3}{c}{$\mathrm{\%MAE}_{\text{weight}}$} & \multicolumn{3}{c}{$\mathrm{\%MAE}_{\text{energy}}$} & \multicolumn{3}{c}{$\mathrm{\%MAE}_{\text{micro/CHO}}$} \\
\cmidrule(lr){4-6}\cmidrule(lr){7-9}\cmidrule(lr){10-12}\cmidrule(lr){13-15}\cmidrule(lr){16-18}\cmidrule(lr){19-21}
\textbf{Query set} & \textbf{Model} & \textbf{Encoder} & $\tau{=}0$ & Def. & $K{=}10$ & $\tau{=}0$ & Def. & $K{=}10$ & $\tau{=}0$ & Def. & $K{=}10$ & $\tau{=}0$ & Def. & $K{=}10$ & $\tau{=}0$ & Def. & $K{=}10$ & $\tau{=}0$ & Def. & $K{=}10$ \\
\midrule
\multirow{12}{*}{Nutrition5k} & \multirow{3}{*}{Gemma-4-31B} & PE-Core & 68.6 & \textbf{64.0} & 68.4 & 53.7 & \textbf{50.1} & 53.0 & \textbf{60.1} & 60.4 & 61.8 & 66.9$^{\blacktriangledown}$ & \textbf{50.7} & 70.5 & 52.8$^{\blacktriangledown}$ & \textbf{44.0} & 57.2 & 57.4$^{\blacktriangledown}$ & \textbf{50.0} & 65.4 \\
& & SigLIP2 & 70.8$^{\blacktriangledown}$ & \textbf{45.7} & 75.7 & 54.2$^{\blacktriangledown}$ & \textbf{44.8} & 60.4 & 61.1$^{\blacktriangledown}$ & \textbf{49.8} & 73.1$^{\blacktriangledown}$ & 64.0$^{\blacktriangledown}$ & \textbf{47.9} & 53.3 & 55.3$^{\blacktriangledown}$ & \textbf{44.3} & 50.5 & 58.8$^{\blacktriangledown}$ & \textbf{48.4} & 55.4 \\
& & Nomic Vision & 66.6 & \textbf{64.4} & 66.7 & 79.7 & \textbf{72.2} & 103.9 & 87.8 & \textbf{80.7} & 102.8 & 70.5$^{\blacktriangledown}$ & \textbf{52.5} & 64.9 & 75.0 & \textbf{55.7} & 70.4 & 85.5 & \textbf{62.5} & 74.5 \\
\cmidrule(lr){2-21}
& \multirow{3}{*}{GPT-5.4-mini} & PE-Core & \textbf{31.1} & 38.5 & 34.9 & \textbf{28.6} & 32.8 & 32.8 & \textbf{32.9} & 40.0 & 36.1 & 39.3 & 35.8 & \textbf{31.1} & 34.0 & 34.7 & \textbf{30.6} & 41.2 & 40.4 & \textbf{34.4} \\
& & SigLIP2 & 33.1 & 33.1 & \textbf{31.4} & 31.4 & \textbf{30.2} & 32.2 & 36.0 & \textbf{35.6} & 38.1 & \textbf{31.8} & 34.9 & 33.0 & 36.3 & 36.6 & \textbf{32.9} & 38.6 & 39.9 & \textbf{37.9} \\
& & Nomic Vision & \textbf{28.2} & 31.1 & 40.5 & 30.6 & \textbf{29.9} & 40.5 & \textbf{37.4} & 37.6 & 47.7 & \textbf{33.1} & 38.2 & 37.2 & \textbf{36.6} & 40.1 & 41.6 & \textbf{45.9} & 46.8 & 48.6 \\
\cmidrule(lr){2-21}
& \multirow{3}{*}{Mistral-Small-3.2} & PE-Core & 279.4 & \textbf{273.4} & 700.8$^{\blacktriangledown}$ & 253.3 & \textbf{244.8} & 649.0$^{\blacktriangledown}$ & 264.4 & \textbf{259.9} & 656.8$^{\blacktriangledown}$ & \textbf{78.6} & \textbf{78.6} & 95.1 & \textbf{70.0} & \textbf{70.0} & 79.5 & \textbf{82.3} & \textbf{82.3} & 99.0 \\
& & SigLIP2 & \textbf{327.9} & 337.1 & 823.8$^{\blacktriangledown}$ & \textbf{281.0} & 283.1 & 761.8$^{\blacktriangledown}$ & \textbf{300.4} & 306.5 & 773.0$^{\blacktriangledown}$ & 89.2 & 89.2 & \textbf{72.3} & 78.5 & 78.5 & \textbf{64.8} & 86.6 & 86.6 & \textbf{75.1} \\
& & Nomic Vision & \textbf{364.4} & 373.3 & 914.2$^{\blacktriangledown}$ & \textbf{394.1} & 418.6 & 989.3$^{\blacktriangledown}$ & \textbf{387.4} & 410.9 & 972.7$^{\blacktriangledown}$ & 102.5 & 104.5 & \textbf{98.3} & 117.9 & \textbf{117.6} & 139.8 & \textbf{123.5} & 124.0 & 149.0 \\
\cmidrule(lr){2-21}
& \multirow{3}{*}{GLM-5.3-Flash} & PE-Core & 32.9 & 55.0 & \textbf{29.8} & 30.7 & 50.8 & \textbf{30.5} & 33.1 & 52.5 & \textbf{32.5} & \textbf{39.3} & 43.5 & 44.6 & 34.9 & 38.4 & \textbf{34.7} & \textbf{36.0} & 39.8 & 37.3 \\
& & SigLIP2 & \textbf{31.5} & 46.5 & 50.9 & \textbf{32.8} & 36.5 & 36.6 & \textbf{34.8} & 40.7 & 41.3 & 39.5 & \textbf{34.1} & 40.2 & 31.8 & \textbf{30.5} & 32.8 & 34.7 & \textbf{33.5} & 35.7 \\
& & Nomic Vision & 39.3 & \textbf{37.8} & 61.5 & 39.6 & \textbf{36.3} & 76.2 & 42.1 & \textbf{42.0} & 80.9 & \textbf{35.6} & 38.1 & 76.8 & 37.4 & \textbf{36.2} & 90.2 & \textbf{40.8} & 40.9 & 95.4 \\
\midrule
\multirow{12}{*}{SNAPMe} & \multirow{3}{*}{Gemma-4-31B} & PE-Core & -- & -- & -- & \textbf{36.0} & 38.8 & 37.5 & \textbf{42.6} & 44.2 & 45.7 & -- & -- & -- & 39.5 & \textbf{38.1} & 38.8 & 44.4 & 42.5 & \textbf{42.3} \\
& & SigLIP2 & -- & -- & -- & 42.4 & \textbf{37.9} & 40.9 & 49.0 & \textbf{44.4} & 49.1 & -- & -- & -- & 39.8 & 36.8 & \textbf{35.6} & 44.2 & 41.2 & \textbf{39.8} \\
& & Nomic Vision & -- & -- & -- & \textbf{36.9} & 39.3 & 42.2 & \textbf{44.5} & 46.4 & 49.1 & -- & -- & -- & 41.0 & 36.3 & \textbf{36.0} & 47.0$^{\blacktriangledown}$ & 41.5 & \textbf{40.7} \\
\cmidrule(lr){2-21}
& \multirow{3}{*}{GPT-5.4-mini} & PE-Core & -- & -- & -- & 37.7 & 34.5 & \textbf{33.4} & 44.0 & 42.0 & \textbf{41.4} & -- & -- & -- & 36.0 & \textbf{33.0} & 33.1 & 42.3 & 41.3 & \textbf{40.8} \\
& & SigLIP2 & -- & -- & -- & 34.5 & 34.5 & \textbf{34.3} & 43.6 & \textbf{42.4} & 43.3 & -- & -- & -- & 34.6 & \textbf{33.6} & 36.0 & 42.4 & \textbf{41.5} & 42.3 \\
& & Nomic Vision & -- & -- & -- & \textbf{34.0} & 36.1 & 35.4 & \textbf{40.8} & 42.4 & 42.7 & -- & -- & -- & 37.7 & 36.3 & \textbf{35.4} & 44.7 & \textbf{43.1} & 43.4 \\
\cmidrule(lr){2-21}
& \multirow{3}{*}{Mistral-Small-3.2} & PE-Core & -- & -- & -- & 118.4 & \textbf{90.9} & 140.3 & 126.4 & \textbf{99.3} & 150.9 & -- & -- & -- & 71.1 & 57.5 & \textbf{48.9} & 83.1$^{\blacktriangledown}$ & 66.6 & \textbf{57.9} \\
& & SigLIP2 & -- & -- & -- & 134.4 & \textbf{126.5} & 202.2 & 140.8 & \textbf{133.7} & 208.7 & -- & -- & -- & 71.9 & 66.3 & \textbf{48.2} & 81.6 & 76.0 & \textbf{56.7} \\
& & Nomic Vision & -- & -- & -- & 110.1 & \textbf{108.0} & 175.4 & 114.6 & \textbf{113.2} & 181.6 & -- & -- & -- & 62.1 & 59.5 & \textbf{44.4} & 69.6 & 67.4 & \textbf{52.3} \\
\cmidrule(lr){2-21}
& \multirow{3}{*}{GLM-5.3-Flash} & PE-Core & -- & -- & -- & 37.7 & \textbf{32.5} & 34.9 & 42.7 & \textbf{38.0} & 39.0 & -- & -- & -- & 32.6 & \textbf{30.8} & 32.5 & 38.5 & \textbf{35.8} & 37.1 \\
& & SigLIP2 & -- & -- & -- & \textbf{34.2} & 35.8 & 34.3 & 40.0 & 41.9 & \textbf{39.1} & -- & -- & -- & 33.2 & 33.7 & \textbf{32.7} & \textbf{38.3} & 39.0 & 38.4 \\
& & Nomic Vision & -- & -- & -- & 37.5 & \textbf{35.5} & 64.8 & 41.8 & \textbf{40.2} & 69.4 & -- & -- & -- & \textbf{32.2} & 32.4 & 37.8 & \textbf{36.7} & 38.5 & 41.5 \\
\midrule
\multirow{12}{*}{MADiMa} & \multirow{3}{*}{Gemma-4-31B} & PE-Core & 40.2 & \textbf{39.5} & 39.9 & 36.6 & 36.7 & \textbf{35.9} & 45.4 & 46.2 & \textbf{45.1} & 74.4 & \textbf{74.3} & 75.1 & 57.1 & 57.4 & \textbf{56.9} & 58.7 & 59.3 & \textbf{58.3} \\
& & SigLIP2 & 38.6 & 37.7 & \textbf{37.5} & 36.1 & 34.8 & \textbf{33.7} & 43.5 & \textbf{42.5} & 43.1 & 74.5 & \textbf{70.3} & 73.1 & 57.9 & 56.6 & \textbf{55.0}$^\star$ & 60.0 & \textbf{58.2} & 58.3 \\
& & Nomic Vision & 44.8 & \textbf{42.4} & 44.5 & 37.8 & \textbf{37.0} & 38.8 & 46.1 & \textbf{45.1} & 46.1 & 81.6 & \textbf{76.9} & 78.8 & 55.5 & \textbf{52.2} & 52.3 & 57.5 & \textbf{55.3} & 56.2 \\
\cmidrule(lr){2-21}
& \multirow{3}{*}{GPT-5.4-mini} & PE-Core & 27.9 & \textbf{26.8} & 28.8 & 33.1 & \textbf{32.3} & 34.3 & 42.7 & \textbf{42.1} & 43.2 & \textbf{40.4} & 43.1 & 40.6 & 39.1 & 37.6 & \textbf{36.5} & 45.2 & 44.6 & \textbf{44.5} \\
& & SigLIP2 & 28.4 & 29.1 & \textbf{26.2} & 33.1 & \textbf{32.0} & 32.6 & 43.5 & 43.1 & \textbf{41.9} & 41.3 & 42.9 & \textbf{35.3}$^\star$ & \textbf{41.8} & 45.2 & 42.2 & 49.2 & 53.0 & \textbf{47.3} \\
& & Nomic Vision & 35.7 & \textbf{31.6} & 32.4 & 36.0 & 33.2 & \textbf{32.3} & 45.9 & \textbf{42.3} & 44.0 & 54.7 & 58.2 & \textbf{54.2} & 40.0 & 41.8 & \textbf{34.9} & 48.0 & 47.9 & \textbf{45.0} \\
\cmidrule(lr){2-21}
& \multirow{3}{*}{Mistral-Small-3.2} & PE-Core & \textbf{27.5}$^\star$ & 52.2 & 31.6 & \textbf{39.9} & 48.6 & 42.2 & \textbf{55.3} & 63.5 & 56.1 & \textbf{30.7}$^\star$ & 34.8 & 31.3 & 38.2 & 40.7 & \textbf{38.1} & \textbf{48.0} & 49.5 & 48.1 \\
& & SigLIP2 & \textbf{34.2}$^\star$ & 62.7 & 38.6 & \textbf{48.0} & 59.8 & 52.5 & \textbf{59.4} & 74.1 & 61.7 & 31.9 & 34.1 & \textbf{28.2} & 40.8 & \textbf{38.7} & 40.7 & 51.0 & 54.0 & \textbf{49.6} \\
& & Nomic Vision & 62.2 & 59.0 & \textbf{57.1} & 67.5 & \textbf{53.6} & 69.6 & 86.7$^{\blacktriangledown}$ & \textbf{67.5} & 83.5 & 44.2 & 43.5 & \textbf{38.2} & 41.1 & 50.7 & \textbf{38.9} & 54.4 & 66.0 & \textbf{52.2} \\
\cmidrule(lr){2-21}
& \multirow{3}{*}{GLM-5.3-Flash} & PE-Core & 31.2 & 32.6 & \textbf{31.1} & 30.4 & 30.4 & \textbf{30.1} & 37.9 & 37.3 & \textbf{37.2} & \textbf{31.7} & 33.6 & 33.4 & 29.2 & 29.8 & \textbf{28.9} & 37.5 & 37.9 & \textbf{36.4} \\
& & SigLIP2 & 32.6 & 32.0 & \textbf{31.8} & 31.6 & 29.5 & \textbf{29.3} & 38.2 & 38.4 & \textbf{37.9} & \textbf{32.0} & 32.4 & 34.0 & \textbf{29.4} & 29.6 & 30.3 & 37.8 & \textbf{36.9} & 37.4 \\
& & Nomic Vision & 37.4 & \textbf{34.9} & 37.5 & \textbf{31.2} & 31.7 & 32.1 & 39.4 & \textbf{37.6} & 39.5 & 38.0 & 38.3 & \textbf{36.6} & 31.7 & \textbf{31.2} & 32.8 & \textbf{37.1} & 37.5 & 38.6 \\
\midrule
\multirow{12}{*}{EDUCATE} & \multirow{3}{*}{Gemma-4-31B} & PE-Core & -- & -- & -- & -- & -- & -- & 38.8 & \textbf{38.5} & 42.4 & -- & -- & -- & -- & -- & -- & \textbf{40.1} & 46.8 & 44.9 \\
& & SigLIP2 & -- & -- & -- & -- & -- & -- & 43.3 & \textbf{41.9} & 42.8 & -- & -- & -- & -- & -- & -- & 47.6 & 48.3 & \textbf{46.9} \\
& & Nomic Vision & -- & -- & -- & -- & -- & -- & 41.2 & 40.4 & \textbf{39.9} & -- & -- & -- & -- & -- & -- & \textbf{47.5} & 50.5 & 48.4 \\
\cmidrule(lr){2-21}
& \multirow{3}{*}{GPT-5.4-mini} & PE-Core & -- & -- & -- & -- & -- & -- & 41.7 & \textbf{36.1} & 44.0 & -- & -- & -- & -- & -- & -- & 49.9 & \textbf{41.1} & 42.4 \\
& & SigLIP2 & -- & -- & -- & -- & -- & -- & 42.4 & \textbf{38.0} & 40.4 & -- & -- & -- & -- & -- & -- & 44.6 & 41.6 & \textbf{41.0} \\
& & Nomic Vision & -- & -- & -- & -- & -- & -- & 42.1 & \textbf{41.9} & 44.4 & -- & -- & -- & -- & -- & -- & 54.7 & \textbf{42.9} & 49.5 \\
\cmidrule(lr){2-21}
& \multirow{3}{*}{Mistral-Small-3.2} & PE-Core & -- & -- & -- & -- & -- & -- & 172.2 & \textbf{118.0} & 181.5 & -- & -- & -- & -- & -- & -- & 112.7 & 91.0 & \textbf{54.7}$^\star$ \\
& & SigLIP2 & -- & -- & -- & -- & -- & -- & 208.3$^{\blacktriangledown}$ & \textbf{134.5} & 174.4 & -- & -- & -- & -- & -- & -- & 115.3 & 96.8 & \textbf{69.7} \\
& & Nomic Vision & -- & -- & -- & -- & -- & -- & 161.9 & \textbf{140.3} & 193.6 & -- & -- & -- & -- & -- & -- & 101.3 & 90.7 & \textbf{71.8} \\
\cmidrule(lr){2-21}
& \multirow{3}{*}{GLM-5.3-Flash} & PE-Core & -- & -- & -- & -- & -- & -- & 54.0 & \textbf{51.9} & 56.4 & -- & -- & -- & -- & -- & -- & 47.7 & \textbf{46.7} & 52.9 \\
& & SigLIP2 & -- & -- & -- & -- & -- & -- & 53.3 & \textbf{52.5} & 55.3 & -- & -- & -- & -- & -- & -- & 46.1 & 46.2 & \textbf{45.2} \\
& & Nomic Vision & -- & -- & -- & -- & -- & -- & \textbf{53.5} & 55.1 & 62.0 & -- & -- & -- & -- & -- & -- & 45.0 & 44.0 & \textbf{43.9} \\
\bottomrule
\end{tabular}}
\end{table*}

\subsubsection{Exploratory Retrieval Extensions}

For each post-hoc food-composition matching method, we evaluate 48 configurations (4 query sets $\times$ 4 VLMs $\times$ 3 visual encoders) using per-item predictions.
Post-hoc food-composition grounding increases $\mathrm{\%MAE}_{\text{micro/CHO}}$ relative to the default NutriVisRAG predictions.
Lexical matching, EmbeddingGemma, and Nomic Embed Text v2 increase the error in all 48 configurations, while Qwen3-Embedding increases it in 46/48.
For energy, which is available for three query sets, the original NutriVisRAG predictions also yield lower error in 34/36 configurations (3 query sets $\times$ 4 VLMs $\times$ 3 visual encoders).
No matching method yields a significant improvement, while significant degradations occur in 45/48 configurations for lexical matching, 30/48 for Qwen3-Embedding, 47/48 for EmbeddingGemma, and 42/48 for Nomic Embed Text v2.
Because these stages were not evaluated independently, the degradations may reflect cumulative pipeline errors rather than inadequate matching models.

For reranking, each prompting strategy comprises 144 comparisons (4 query sets $\times$ 4 VLMs $\times$ 3 visual encoders $\times$ 3 rerankers), corresponding to 36 comparisons per query set.
Under per-item prompting, reranking reduces $\mathrm{\%MAE}_{\text{micro/CHO}}$ in 73/144 comparisons, including 24/36 on MADiMa, 19/36 on Nutrition5k, 17/36 on EDUCATE, and 13/36 on SNAPMe.
Eight per-item improvements and two degradations remain significant.
Under whole-meal prompting, reranking reduces error in 76/144 comparisons, including 28/36 on MADiMa, 21/36 on EDUCATE, 17/36 on SNAPMe, and 10/36 on Nutrition5k.
Six whole-meal improvements and four degradations remain significant after correction.
Reranking can therefore recover more useful exemplars in some configurations, but the rerankers optimise general multimodal relevance rather than nutritional equivalence, so a semantically more relevant meal is not necessarily a better quantitative reference.
Reranking additionally adds a median of 5.3--11.7\,s per meal, increasing end-to-end latency by $2.6\times$--$5.1\times$ depending on the reranker and VLM.
Overall, its variable accuracy gains and additional computational cost do not support incorporating reranking into the default NutriVisRAG pipeline.

\subsection{Limitations and Future Work}
\label{subsec:limitations}

The findings remain specific to the evaluated VLMs, retrieval encoders, and combined reference corpus, whose coverage may not represent the diversity of real-world meals and whose annotations combine measured Nutrition5k quantities with participant-reported SNAPMe quantities.
Broader evaluation across VLM families, visual representations, and larger, more diverse reference corpora is needed, particularly as MADiMa and EDUCATE remain entirely outside the indexed corpus and EDUCATE is evaluated only for CHO against dietitians' visual estimates rather than measured ground truth.
Ablations were conducted on fixed 100-query subsets rather than the full query sets, and their observed effects may not generalise to complete datasets.
Matched textual-retrieval and random-exemplar controls were not evaluated, limiting attribution of the observed gains specifically to visual similarity-based grounding.
We also fix the VLM input image size; future work should evaluate query and exemplar resolution independently to determine whether finer visual detail justifies greater visual-token usage and inference latency.
Our similarity analysis indicates that visual similarity is an imperfect proxy for exemplar usefulness, motivating more adaptive retrieval or reranking criteria that capture ingredient, portion, compositional, and quantitative relevance using information available at inference.

\section{Conclusion}
\label{sec:conclusion}
We introduced NutriVisRAG, an inference-time framework that grounds pretrained VLMs with visually retrieved meals and their nutritional annotations as multimodal quantitative exemplars, without task-specific fine-tuning.
Across four dietary datasets, four VLMs, three retrieval encoders, and two prompting strategies, the lowest observed NutriVisRAG error was below the best Vanilla error in 13 of 16 VLM--dataset pairs.
More importantly, our investigation shows that retrieval effectiveness is configuration-dependent and cannot be explained by visual similarity alone: benefits vary across VLMs and prompting strategies, while visually closer exemplars do not consistently correspond to greater downstream improvement or nutritional proximity.
The more consistent gains under whole-meal prompting further suggest that exemplar usefulness depends not only on what is retrieved, but also on how the retrieved evidence is presented and used by the VLM.
Retrieval ablations reinforce this interpretation, showing that similarity filtering is often beneficial while increasing retrieval depth provides no consistent accuracy benefit, and additional post-hoc grounding and reranking provide no consistently better alternative.
Future work should therefore investigate inference-time retrieval or reranking criteria that better capture exemplar usefulness without requiring query nutrient annotations, alongside finer-grained analyses of query--exemplar correspondence and evaluation using larger and more diverse reference corpora.


\textbf{Acknowledgements}
This work was partly funded by the European Commission and the Swiss State Secretariat for Education, Research and Innovation (SERI) within the MELISSA project (101057730) and BETTER4U (101080117).




\bibliographystyle{splncs04}
\bibliography{main}

\setcounter{section}{0}
\renewcommand{\thesection}{S\arabic{section}}
\renewcommand{\thesubsection}{\thesection.\arabic{subsection}}
\clearpage
\begin{center}
  {\Large\bfseries Supplementary Material: NutriVisRAG\par}
\end{center}
\vspace{1em}


\setcounter{section}{0}
\renewcommand{\thesection}{S\arabic{section}}
\renewcommand{\thesubsection}{\thesection.\arabic{subsection}}

\setcounter{figure}{0}
\renewcommand{\thefigure}{S\arabic{figure}}

\setcounter{table}{0}
\renewcommand{\thetable}{S\arabic{table}}

\section{Exact Prompt Templates}
\label{sec:prompt-templates}

This section reproduces the textual prompt templates used for Vanilla and NutriVisRAG inference.
Angle-bracketed elements indicate query-dependent images or annotations inserted programmatically and are not literal text tokens.
For NutriVisRAG, the retrieved exemplars are inserted in descending similarity order before the query image.
The prompt wording is reproduced verbatim; only the typesetting is adapted for readability.

\subsection{Vanilla VLM Inference}

\begin{promptbox}{Vanilla \texttt{per-item}}
\small\raggedright
\textbf{System}\par\smallskip

You are a meal image nutritional analysis tool.\par
You are given a user QUERY meal image.\par
Estimate the item-level weight, energy, and macronutrients\par
To do this, think step by step as follows:\par
1. Detect all visually distinguishable edible food items using visual cues such as colour, texture, shape, and spatial boundaries.\par
2. Give each detected food item the most specific food label the image visually supports, as lowercase words separated by underscores (e.g., mashed\_potatoes, beef\_stroganoff, chicken\_rice\_pilaf).\par
3. Estimate the weight (in grams), energy (in kcal), carbohydrates (in grams), protein (in grams), and fat (in grams) for each detected food item.\par
4. Assign each food item a confidence score between 0 and 1 reflecting your overall confidence in the food identification and nutritional estimate.\par
5. For each food item, briefly explain the visual evidence supporting the food identification and nutritional estimate.\par
Return structured JSON only --- no prose, no markdown.

\medskip
\tcbline
\medskip

\textbf{User}\par\smallskip

$\langle$\textit{query image}$\rangle$\par
For this meal image, identify each visible food item and estimate its weight (in grams), energy (in kcal), carbohydrates (in grams), protein (in grams) and fat (in grams), and give each a confidence score (0-1) and a brief justification of the visual evidence.
\end{promptbox}

\begin{promptbox}{Vanilla \texttt{whole-meal}}
\small\raggedright
\textbf{System}\par\smallskip

You are a meal image nutritional analysis tool.\par
You are given a user QUERY meal image.\par
Estimate the total weight of the whole meal in grams, then the whole-meal totals for energy (in kcal), carbohydrates (in grams), protein (in grams), and fat (in grams).\par
Provide an overall confidence score between 0 and 1 and a brief justification of the visual evidence.\par
Return structured JSON only --- no prose, no markdown.

\medskip
\tcbline
\medskip

\textbf{User}\par\smallskip

$\langle$\textit{query image}$\rangle$\par
For this meal image, estimate the total meal weight (in grams), then the total energy (in kcal), carbohydrates (in grams), protein (in grams) and fat (in grams), give an overall confidence score (0-1) and a brief justification of the visual evidence.
\end{promptbox}

\subsection{NutriVisRAG Inference}

The main experiments retrieve from the combined Nutrition5k and SNAPMe reference corpus.
Because all retrieved exemplars contain energy and macronutrient annotations but measured weight is available only for Nutrition5k exemplars, the system prompt explicitly communicates this distinction to the VLM.

\begin{promptbox}{NutriVisRAG \texttt{per-item}}
\small\raggedright
\textbf{System}\par\smallskip

You are a meal image nutritional analysis tool.\par
You are given a user QUERY meal image and reference images, each labeled with energy (in kcal), carbohydrates (in grams), protein (in grams), and fat (in grams) at meal and meal-item level.\par
Additionally, in certain cases, weight (in grams) of meal and meal-item level is given to you.\par
Use the provided reference information as guidance, whenever possible, to estimate the item-level weight, energy, and macronutrients. To do this, think step by step as follows:\par
1. Detect all visually distinguishable edible food items using visual cues such as colour, texture, shape, and spatial boundaries.\par
2. Give each detected food item the most specific food label the image visually supports, as lowercase words separated by underscores (e.g., mashed\_potatoes, beef\_stroganoff, chicken\_rice\_pilaf).\par
3. Estimate the weight (in grams), energy (in kcal), carbohydrates (in grams), protein (in grams), and fat (in grams) for each detected food item.\par
4. Assign each food item a confidence score between 0 and 1 reflecting your overall confidence in the food identification and nutritional estimate.\par
5. For each food item, briefly explain the visual evidence supporting the food identification and nutritional estimate.\par
Return structured JSON only --- no prose, no markdown.

\medskip
\tcbline
\medskip

\textbf{User}\par\smallskip

Here are similar dishes for your reference to help you with the estimations:\par\smallskip
$\langle$\textit{reference image 1}$\rangle$\par
$\langle$\textit{reference 1 nutritional annotations}$\rangle$\par\smallskip
$\langle$\textit{reference image 2}$\rangle$\par
$\langle$\textit{reference 2 nutritional annotations}$\rangle$\par\smallskip
$\vdots$\par\smallskip
$\langle$\textit{reference image $K$}$\rangle$\par
$\langle$\textit{reference $K$ nutritional annotations}$\rangle$\par\medskip
$\langle$\textit{query image}$\rangle$\par
For this meal image, identify each visible food item and estimate its weight (in grams), energy (in kcal), carbohydrates (in grams), protein (in grams) and fat (in grams), and give each a confidence score (0-1) and a brief justification of the visual evidence.
\end{promptbox}

\begin{promptbox}{NutriVisRAG \texttt{whole-meal}}
\small\raggedright
\textbf{System}\par\smallskip

You are a meal image nutritional analysis tool.\par
You are given a user QUERY meal image and reference images, each labeled with energy (in kcal), carbohydrates (in grams), protein (in grams), and fat (in grams) at meal and meal-item level.\par
Additionally, in certain cases, weight (in grams) of meal and meal-item level is given to you.\par
Use the provided reference information as guidance, whenever possible, to estimate the total weight of the whole meal in grams, then the whole-meal totals for energy (in kcal), carbohydrates (in grams), protein (in grams), and fat (in grams).\par
Provide an overall confidence score between 0 and 1 and a brief justification of the visual evidence.\par
Return structured JSON only --- no prose, no markdown.

\medskip
\tcbline
\medskip

\textbf{User}\par\smallskip

Here are similar dishes for your reference to help you with the estimations:\par\smallskip
$\langle$\textit{reference image 1}$\rangle$\par
$\langle$\textit{reference 1 nutritional annotations}$\rangle$\par\smallskip
$\langle$\textit{reference image 2}$\rangle$\par
$\langle$\textit{reference 2 nutritional annotations}$\rangle$\par\smallskip
$\vdots$\par\smallskip
$\langle$\textit{reference image $K$}$\rangle$\par
$\langle$\textit{reference $K$ nutritional annotations}$\rangle$\par\medskip
$\langle$\textit{query image}$\rangle$\par
For this meal image, estimate the total meal weight (in grams), then the total energy (in kcal), carbohydrates (in grams), protein (in grams) and fat (in grams), give an overall confidence score (0-1) and a brief justification of the visual evidence.
\end{promptbox}

\subsection{Retrieved-Exemplar Annotation Format}

The nutritional annotation inserted after each retrieved image depends on the source dataset.
Measured weight is included for Nutrition5k exemplars and omitted for SNAPMe exemplars rather than estimated or imputed.

\begin{promptbox}{Nutrition5k exemplar}
\small\raggedright
Reference $\langle j\rangle$ --- Items: $\langle$\textit{item-level annotations including weight, energy, carbohydrates, protein, and fat}$\rangle$. Meal total: $\langle$\textit{weight, energy, carbohydrates, protein, and fat}$\rangle$.
\end{promptbox}

\begin{promptbox}{SNAPMe exemplar}
\small\raggedright
Reference $\langle j\rangle$ --- Items: $\langle$\textit{item-level annotations including energy, carbohydrates, protein, and fat}$\rangle$. Meal total: $\langle$\textit{energy, carbohydrates, protein, and fat}$\rangle$.
\end{promptbox}

\section{Effect of Acquisition Distance on MADiMa in Vanilla-Inference}
\label{sec:direct-additional}

\begin{figure}[tb]
\centering
\includegraphics[width=\textwidth]{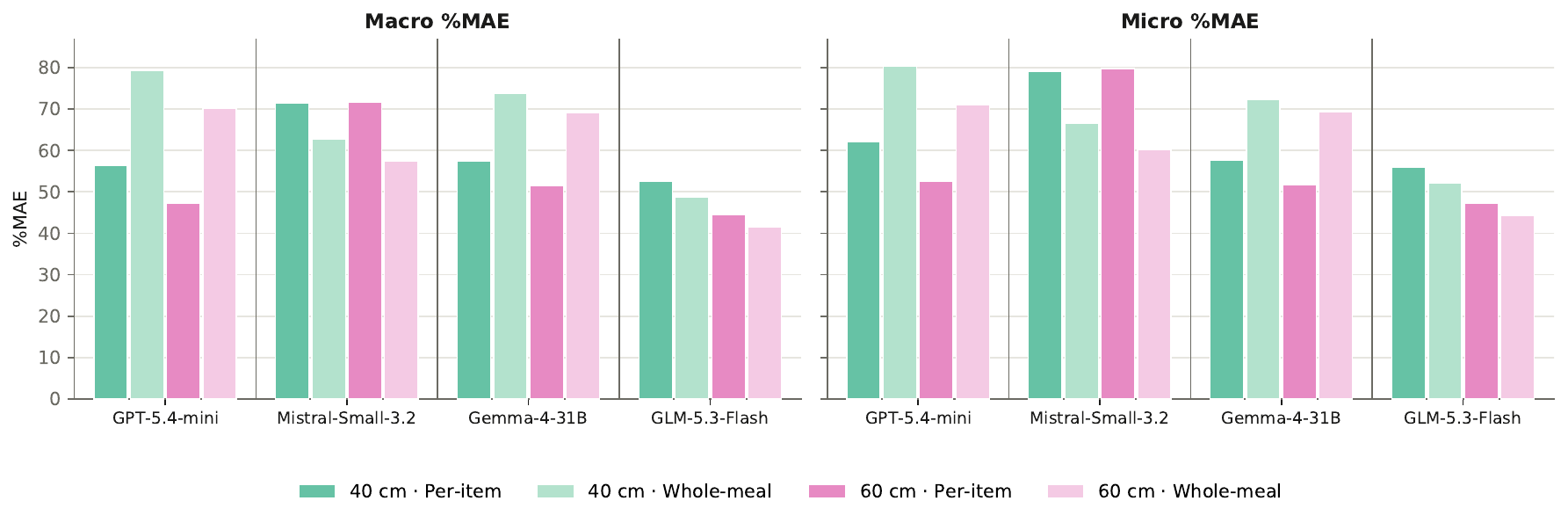}
\caption{Effect of acquisition distance on Vanilla VLM inference for MADiMa.
$\mathrm{\%MAE}_{\text{macro}}$ and $\mathrm{\%MAE}_{\text{micro}}$ (lower is better) are shown at 40\,cm and 60\,cm for per-item and whole-meal prompting across the four evaluated VLMs.}
\label{fig:madima-distance}
\end{figure}

MADiMa provides meal images acquired at both 40\,cm and 60\,cm, allowing performance to be compared across acquisition distances.
Figure~\ref{fig:madima-distance} reports $\mathrm{\%MAE}_{\text{macro}}$ and $\mathrm{\%MAE}_{\text{micro}}$ for the two prompting strategies across the four evaluated VLMs.
The 60\,cm images yield lower $\mathrm{\%MAE}_{\text{macro}}$ and $\mathrm{\%MAE}_{\text{micro}}$ for GPT-5.4-mini, Gemma-4-31B, and GLM-5.3-Flash under both prompting strategies, and for Mistral-Small-3.2 under whole-meal prompting.
No corresponding improvement is observed for Mistral-Small-3.2 with per-item prompting.

\section{Complete NutriVisRAG Results}
\label{sec:complete-rag}

Unless otherwise stated, \%MAE values in the supplementary tables and accompanying text are rounded to one decimal place.
Table~\ref{tab:rag-results-full} reports the complete NutriVisRAG results for both prompting strategies and all evaluated retrieval encoders.
Table~4 of the main paper reports the primary comparison metric ($\mathrm{\%MAE}_{\text{micro}}$ for Nutrition5k, SNAPMe, and MADiMa; $\mathrm{\%MAE}_{\text{CHO}}$ for EDUCATE), while Table~\ref{tab:rag-results-full} additionally reports the error for each available nutritional target.

\begin{table}[tb]
\centering
\caption{
Complete nutrient-estimation results with and without NutriVisRAG (\%MAE, lower is better), for both prompting strategies side by side.
Dashes indicate unavailable evaluation targets for the corresponding dataset.
Bold: lowest Macro and Micro (or CHO) error within each model block and prompting strategy; arrows: increase ($\uparrow$) or decrease ($\downarrow$) relative to the Vanilla baseline of the same prompting strategy.
$^\star$ marks a statistically significant improvement over that baseline and $^{\blacktriangledown}$ a statistically significant degradation (paired Wilcoxon signed-rank test on per-meal absolute errors, Benjamini--Hochberg corrected, $q<0.05$); the Macro column is an average of per-nutrient errors and is therefore not tested. Arrows indicate the direction of change based on unrounded values; displayed values are rounded to one decimal place.
}
\label{tab:rag-results-full}
\scriptsize
\setlength{\tabcolsep}{1.5pt}
\renewcommand{\arraystretch}{0.92}

\resizebox{\textwidth}{!}{%
\begin{tabular}{lllrrrrrrrrrrrrrr}
\toprule
& & & \multicolumn{7}{c}{\textbf{Per-item}} & \multicolumn{7}{c}{\textbf{Whole-meal}} \\
\cmidrule(lr){4-10}\cmidrule(lr){11-17}
\textbf{Query set} & \textbf{Model} & \textbf{Setting} & \textbf{Weight} & \textbf{Energy} & \textbf{CHO} & \textbf{Protein} & \textbf{Fat} & \textbf{Macro} & \textbf{Micro} & \textbf{Weight} & \textbf{Energy} & \textbf{CHO} & \textbf{Protein} & \textbf{Fat} & \textbf{Macro} & \textbf{Micro} \\
\midrule
Nutrition5k & Gemma-4-31B & Vanilla & 36.3 & 47.7 & 56.0 & 44.3 & 63.3 & 54.5 & \textbf{53.6} & 49.2 & 58.4 & 69.6 & 46.4 & 78.4 & 64.8 & 63.5 \\
 &  & NutriVisRAG (PE-Core) & 48.4$\uparrow$$^{\blacktriangledown}$ & 46.0$\downarrow$ & 61.0$\uparrow$ & 45.8$\uparrow$ & 52.9$\downarrow$$^\star$ & \textbf{53.2}$\downarrow$ & 53.7$\uparrow$ & 47.4$\downarrow$$^\star$ & 44.9$\downarrow$$^\star$ & 55.9$\downarrow$$^\star$ & 43.9$\downarrow$$^\star$ & 53.4$\downarrow$$^\star$ & \textbf{51.0}$\downarrow$ & 51.0$\downarrow$$^\star$ \\
 &  & NutriVisRAG (SigLIP2) & 47.6$\uparrow$$^{\blacktriangledown}$ & 47.8$\uparrow$ & 58.6$\uparrow$ & 47.1$\uparrow$ & 58.8$\downarrow$$^\star$ & 54.9$\uparrow$ & 54.6$\uparrow$ & 46.7$\downarrow$$^\star$ & 45.3$\downarrow$$^\star$ & 54.6$\downarrow$$^\star$ & 43.5$\downarrow$$^\star$ & 55.1$\downarrow$$^\star$ & 51.1$\downarrow$ & \textbf{50.8}$\downarrow$$^\star$ \\
 &  & NutriVisRAG (Nomic Vision) & 46.3$\uparrow$$^{\blacktriangledown}$ & 51.4$\uparrow$ & 59.8$\uparrow$ & 55.5$\uparrow$ & 66.0$\uparrow$ & 60.5$\uparrow$ & 59.8$\uparrow$ & 46.5$\downarrow$$^\star$ & 49.6$\downarrow$$^\star$ & 56.5$\downarrow$$^\star$ & 51.2$\uparrow$ & 61.4$\downarrow$$^\star$ & 56.4$\downarrow$ & 55.8$\downarrow$$^\star$ \\
\midrule
 & GPT-5.4-mini & Vanilla & 35.8 & 47.2 & 66.2 & 45.8 & 62.2 & 58.0 & 58.0 & 40.1 & 48.1 & 60.8 & 47.2 & 69.9 & 59.3 & 58.1 \\
 &  & NutriVisRAG (PE-Core) & 31.7$\downarrow$$^\star$ & 36.7$\downarrow$$^\star$ & 44.3$\downarrow$$^\star$ & 37.7$\downarrow$$^\star$ & 47.4$\downarrow$$^\star$ & \textbf{43.1}$\downarrow$ & \textbf{42.7}$\downarrow$$^\star$ & 34.2$\downarrow$$^\star$ & 36.3$\downarrow$$^\star$ & 45.7$\downarrow$$^\star$ & 39.0$\downarrow$$^\star$ & 46.0$\downarrow$$^\star$ & \textbf{43.5}$\downarrow$ & \textbf{43.4}$\downarrow$$^\star$ \\
 &  & NutriVisRAG (SigLIP2) & 31.6$\downarrow$$^\star$ & 36.4$\downarrow$$^\star$ & 46.2$\downarrow$$^\star$ & 37.4$\downarrow$$^\star$ & 47.4$\downarrow$$^\star$ & 43.7$\downarrow$ & 43.4$\downarrow$$^\star$ & 34.5$\downarrow$$^\star$ & 36.9$\downarrow$$^\star$ & 46.8$\downarrow$$^\star$ & 40.0$\downarrow$$^\star$ & 46.0$\downarrow$$^\star$ & 44.3$\downarrow$ & 44.2$\downarrow$$^\star$ \\
 &  & NutriVisRAG (Nomic Vision) & 33.7$\downarrow$$^\star$ & 38.8$\downarrow$$^\star$ & 49.2$\downarrow$$^\star$ & 42.2$\downarrow$$^\star$ & 50.9$\downarrow$$^\star$ & 47.4$\downarrow$ & 47.1$\downarrow$$^\star$ & 36.2$\downarrow$$^\star$ & 40.5$\downarrow$$^\star$ & 48.3$\downarrow$$^\star$ & 44.0$\downarrow$$^\star$ & 53.6$\downarrow$$^\star$ & 48.6$\downarrow$ & 48.0$\downarrow$$^\star$ \\
\midrule
 & Mistral-Small-3.2 & Vanilla & 46.0 & 56.2 & 85.4 & 58.8 & 69.3 & \textbf{71.2} & \textbf{72.1} & 62.2 & 60.8 & 97.5 & 60.0 & 74.3 & 77.3 & 78.7 \\
 &  & NutriVisRAG (PE-Core) & 117.8$\uparrow$$^{\blacktriangledown}$ & 99.5$\uparrow$$^{\blacktriangledown}$ & 140.7$\uparrow$$^{\blacktriangledown}$ & 86.0$\uparrow$$^{\blacktriangledown}$ & 98.4$\uparrow$ & 108.4$\uparrow$ & 111.2$\uparrow$$^{\blacktriangledown}$ & 52.9$\downarrow$$^\star$ & 53.1$\downarrow$$^\star$ & 74.3$\downarrow$$^\star$ & 50.4$\downarrow$$^\star$ & 59.8$\downarrow$$^\star$ & \textbf{61.5}$\downarrow$ & \textbf{62.3}$\downarrow$$^\star$ \\
 &  & NutriVisRAG (SigLIP2) & 124.7$\uparrow$$^{\blacktriangledown}$ & 109.7$\uparrow$$^{\blacktriangledown}$ & 144.4$\uparrow$$^{\blacktriangledown}$ & 94.6$\uparrow$$^{\blacktriangledown}$ & 114.1$\uparrow$$^{\blacktriangledown}$ & 117.7$\uparrow$ & 119.5$\uparrow$$^{\blacktriangledown}$ & 53.9$\downarrow$$^\star$ & 55.0$\downarrow$$^\star$ & 72.0$\downarrow$$^\star$ & 51.5$\downarrow$$^\star$ & 66.2$\downarrow$$^\star$ & 63.2$\downarrow$ & 63.3$\downarrow$$^\star$ \\
 &  & NutriVisRAG (Nomic Vision) & 134.3$\uparrow$$^{\blacktriangledown}$ & 137.2$\uparrow$$^{\blacktriangledown}$ & 164.7$\uparrow$$^{\blacktriangledown}$ & 126.0$\uparrow$$^{\blacktriangledown}$ & 140.1$\uparrow$$^{\blacktriangledown}$ & 143.6$\uparrow$ & 145.1$\uparrow$$^{\blacktriangledown}$ & 54.6$\downarrow$$^\star$ & 65.1$\uparrow$ & 79.5$\downarrow$$^\star$ & 65.1$\uparrow$ & 79.3$\uparrow$ & 74.6$\downarrow$ & 74.3$\downarrow$$^\star$ \\
\midrule
 & GLM-5.3-Flash & Vanilla & 38.0 & 45.6 & 62.5 & 40.6 & 59.2 & 54.1 & 54.0 & 42.0 & 47.8 & 65.5 & 41.9 & 61.5 & \textbf{56.3} & \textbf{56.2} \\
 &  & NutriVisRAG (PE-Core) & 52.8$\uparrow$ & 47.2$\uparrow$$^{\blacktriangledown}$ & 61.4$\downarrow$$^\star$ & 48.8$\uparrow$$^{\blacktriangledown}$ & 49.8$\downarrow$$^\star$ & 53.3$\downarrow$ & 54.1$\uparrow$$^{\blacktriangledown}$ & 59.8$\uparrow$$^{\blacktriangledown}$ & 53.6$\uparrow$$^{\blacktriangledown}$ & 62.8$\downarrow$$^\star$ & 55.5$\uparrow$$^{\blacktriangledown}$ & 56.9$\downarrow$$^\star$ & 58.4$\uparrow$ & 58.8$\uparrow$$^{\blacktriangledown}$ \\
 &  & NutriVisRAG (SigLIP2) & 51.9$\uparrow$ & 46.5$\uparrow$$^{\blacktriangledown}$ & 59.4$\downarrow$$^\star$ & 48.5$\uparrow$$^{\blacktriangledown}$ & 50.4$\downarrow$$^\star$ & \textbf{52.8}$\downarrow$ & \textbf{53.4}$\downarrow$$^\star$ & 56.9$\uparrow$$^{\blacktriangledown}$ & 51.1$\uparrow$$^{\blacktriangledown}$ & 61.4$\downarrow$$^\star$ & 52.2$\uparrow$$^{\blacktriangledown}$ & 55.9$\downarrow$$^\star$ & 56.5$\uparrow$ & 56.8$\uparrow$$^{\blacktriangledown}$ \\
 &  & NutriVisRAG (Nomic Vision) & 51.7$\uparrow$ & 53.0$\uparrow$$^{\blacktriangledown}$ & 60.6$\downarrow$$^\star$ & 55.7$\uparrow$ & 59.5$\uparrow$$^{\blacktriangledown}$ & 58.6$\uparrow$ & 58.6$\uparrow$$^{\blacktriangledown}$ & 56.1$\uparrow$$^{\blacktriangledown}$ & 56.7$\uparrow$$^{\blacktriangledown}$ & 62.8$\downarrow$$^\star$ & 61.0$\uparrow$ & 62.8$\uparrow$$^{\blacktriangledown}$ & 62.2$\uparrow$ & 62.2$\uparrow$$^{\blacktriangledown}$ \\
\midrule
SNAPMe & Gemma-4-31B & Vanilla & -- & 35.8 & 38.0 & 39.6 & 48.4 & \textbf{42.0} & \textbf{41.0} & -- & 39.3 & 41.4 & 40.8 & 51.0 & 44.4 & 43.6 \\
 &  & NutriVisRAG (PE-Core) & -- & 41.4$\uparrow$$^{\blacktriangledown}$ & 45.9$\uparrow$$^{\blacktriangledown}$ & 43.2$\uparrow$ & 50.2$\uparrow$ & 46.4$\uparrow$ & 46.2$\uparrow$$^{\blacktriangledown}$ & -- & 37.6$\downarrow$ & 40.4$\downarrow$ & 39.9$\downarrow$ & 49.3$\downarrow$ & 43.2$\downarrow$ & 42.5$\downarrow$ \\
 &  & NutriVisRAG (SigLIP2) & -- & 38.2$\uparrow$ & 41.1$\uparrow$ & 42.9$\uparrow$ & 50.5$\uparrow$ & 44.9$\uparrow$ & 44.0$\uparrow$ & -- & 35.6$\downarrow$$^\star$ & 38.1$\downarrow$$^\star$ & 38.6$\downarrow$ & 46.9$\downarrow$ & \textbf{41.2}$\downarrow$ & \textbf{40.4}$\downarrow$$^\star$ \\
 &  & NutriVisRAG (Nomic Vision) & -- & 38.8$\uparrow$ & 43.5$\uparrow$$^{\blacktriangledown}$ & 43.6$\uparrow$$^{\blacktriangledown}$ & 49.4$\uparrow$ & 45.5$\uparrow$ & 45.0$\uparrow$$^{\blacktriangledown}$ & -- & 36.3$\downarrow$ & 39.6$\downarrow$ & 40.0$\downarrow$ & 48.1$\downarrow$ & 42.6$\downarrow$ & 41.9$\downarrow$ \\
\midrule
 & GPT-5.4-mini & Vanilla & -- & 35.0 & 39.4 & 40.1 & 46.5 & \textbf{42.0} & \textbf{41.4} & -- & 35.5 & 40.1 & 38.9 & 48.0 & 42.3 & 41.7 \\
 &  & NutriVisRAG (PE-Core) & -- & 36.3$\uparrow$ & 38.5$\downarrow$ & 41.1$\uparrow$ & 48.4$\uparrow$ & 42.7$\uparrow$ & 41.7$\uparrow$ & -- & 37.4$\uparrow$ & 40.9$\uparrow$ & 40.7$\uparrow$ & 48.7$\uparrow$ & 43.4$\uparrow$ & 42.8$\uparrow$ \\
 &  & NutriVisRAG (SigLIP2) & -- & 35.8$\uparrow$ & 40.5$\uparrow$ & 41.1$\uparrow$ & 46.6$\uparrow$ & 42.7$\uparrow$ & 42.2$\uparrow$ & -- & 35.1$\downarrow$ & 38.9$\downarrow$ & 39.4$\uparrow$ & 46.6$\downarrow$ & \textbf{41.6}$\downarrow$ & \textbf{41.0}$\downarrow$ \\
 &  & NutriVisRAG (Nomic Vision) & -- & 37.1$\uparrow$ & 39.2$\downarrow$ & 42.4$\uparrow$ & 50.1$\uparrow$$^{\blacktriangledown}$ & 43.9$\uparrow$ & 42.8$\uparrow$ & -- & 36.8$\uparrow$ & 41.7$\uparrow$ & 41.7$\uparrow$ & 46.9$\downarrow$ & 43.4$\uparrow$ & 43.0$\uparrow$ \\
\midrule
 & Mistral-Small-3.2 & Vanilla & -- & 42.6 & 47.8 & 49.8 & 58.6 & \textbf{52.1} & \textbf{51.1} & -- & 37.3 & 41.0 & 43.7 & 52.5 & \textbf{45.7} & \textbf{44.6} \\
 &  & NutriVisRAG (PE-Core) & -- & 93.7$\uparrow$$^{\blacktriangledown}$ & 107.7$\uparrow$$^{\blacktriangledown}$ & 105.3$\uparrow$$^{\blacktriangledown}$ & 94.0$\uparrow$$^{\blacktriangledown}$ & 102.4$\uparrow$ & 103.6$\uparrow$$^{\blacktriangledown}$ & -- & 58.4$\uparrow$$^{\blacktriangledown}$ & 66.1$\uparrow$$^{\blacktriangledown}$ & 64.7$\uparrow$$^{\blacktriangledown}$ & 71.9$\uparrow$$^{\blacktriangledown}$ & 67.6$\uparrow$ & 67.2$\uparrow$$^{\blacktriangledown}$ \\
 &  & NutriVisRAG (SigLIP2) & -- & 120.3$\uparrow$$^{\blacktriangledown}$ & 138.3$\uparrow$$^{\blacktriangledown}$ & 128.4$\uparrow$$^{\blacktriangledown}$ & 117.0$\uparrow$$^{\blacktriangledown}$ & 127.9$\uparrow$ & 130.3$\uparrow$$^{\blacktriangledown}$ & -- & 64.9$\uparrow$$^{\blacktriangledown}$ & 74.1$\uparrow$$^{\blacktriangledown}$ & 72.3$\uparrow$$^{\blacktriangledown}$ & 79.8$\uparrow$$^{\blacktriangledown}$ & 75.4$\uparrow$ & 75.0$\uparrow$$^{\blacktriangledown}$ \\
 &  & NutriVisRAG (Nomic Vision) & -- & 97.5$\uparrow$$^{\blacktriangledown}$ & 118.0$\uparrow$$^{\blacktriangledown}$ & 103.8$\uparrow$$^{\blacktriangledown}$ & 96.8$\uparrow$$^{\blacktriangledown}$ & 106.2$\uparrow$ & 108.8$\uparrow$$^{\blacktriangledown}$ & -- & 57.2$\uparrow$$^{\blacktriangledown}$ & 64.2$\uparrow$$^{\blacktriangledown}$ & 63.9$\uparrow$$^{\blacktriangledown}$ & 71.9$\uparrow$$^{\blacktriangledown}$ & 66.7$\uparrow$ & 66.0$\uparrow$$^{\blacktriangledown}$ \\
\midrule
 & GLM-5.3-Flash & Vanilla & -- & 34.0 & 36.9 & 38.6 & 46.5 & \textbf{40.7} & \textbf{39.8} & -- & 34.6 & 37.9 & 38.1 & 44.7 & 40.2 & 39.6 \\
 &  & NutriVisRAG (PE-Core) & -- & 35.0$\uparrow$ & 37.2$\uparrow$ & 39.4$\uparrow$ & 46.2$\downarrow$ & 41.0$\uparrow$ & 40.1$\uparrow$ & -- & 33.1$\downarrow$$^\star$ & 35.2$\downarrow$$^\star$ & 38.0$\downarrow$ & 43.4$\downarrow$ & \textbf{38.9}$\downarrow$ & \textbf{38.0}$\downarrow$$^\star$ \\
 &  & NutriVisRAG (SigLIP2) & -- & 38.5$\uparrow$ & 40.6$\uparrow$ & 44.5$\uparrow$ & 49.2$\uparrow$ & 44.8$\uparrow$ & 43.8$\uparrow$ & -- & 33.2$\downarrow$ & 36.1$\downarrow$ & 38.2$\uparrow$ & 43.4$\downarrow$ & 39.2$\downarrow$ & 38.5$\downarrow$ \\
 &  & NutriVisRAG (Nomic Vision) & -- & 37.2$\uparrow$ & 38.5$\uparrow$ & 40.2$\uparrow$ & 48.5$\uparrow$ & 42.4$\uparrow$ & 41.5$\uparrow$ & -- & 33.6$\downarrow$ & 35.5$\downarrow$ & 39.0$\uparrow$ & 44.7$\uparrow$ & 39.8$\downarrow$ & 38.8$\downarrow$ \\
\bottomrule
\end{tabular}%
}
\end{table}

\begin{table}[tb]
\centering
\addtocounter{table}{-1}
\caption{(continued) Complete nutrient-estimation results with and without NutriVisRAG.}
\scriptsize
\setlength{\tabcolsep}{1.5pt}
\renewcommand{\arraystretch}{0.92}
\resizebox{\textwidth}{!}{%
\begin{tabular}{lllrrrrrrrrrrrrrr}
\toprule
& & & \multicolumn{7}{c}{\textbf{Per-item}} & \multicolumn{7}{c}{\textbf{Whole-meal}} \\
\cmidrule(lr){4-10}\cmidrule(lr){11-17}
\textbf{Query set} & \textbf{Model} & \textbf{Setting} & \textbf{Weight} & \textbf{Energy} & \textbf{CHO} & \textbf{Protein} & \textbf{Fat} & \textbf{Macro} & \textbf{Micro} & \textbf{Weight} & \textbf{Energy} & \textbf{CHO} & \textbf{Protein} & \textbf{Fat} & \textbf{Macro} & \textbf{Micro} \\
\midrule
MADiMa & Gemma-4-31B & Vanilla & 51.2 & 49.1 & 60.1 & 38.9 & 64.2 & 54.4 & 54.7 & 76.5 & 69.5 & 76.4 & 48.5 & 89.1 & 71.3 & 70.7 \\
 &  & NutriVisRAG (PE-Core) & 38.8$\downarrow$$^\star$ & 40.5$\downarrow$$^\star$ & 54.6$\downarrow$$^\star$ & 41.7$\uparrow$ & 52.7$\downarrow$$^\star$ & 49.7$\downarrow$ & 50.5$\downarrow$$^\star$ & 71.9$\downarrow$$^\star$ & 62.4$\downarrow$$^\star$ & 64.9$\downarrow$$^\star$ & 43.1$\downarrow$$^\star$ & 85.8$\downarrow$ & 64.6$\downarrow$ & 62.7$\downarrow$$^\star$ \\
 &  & NutriVisRAG (SigLIP2) & 38.3$\downarrow$$^\star$ & 38.8$\downarrow$$^\star$ & 51.1$\downarrow$$^\star$ & 37.5$\downarrow$ & 53.3$\downarrow$$^\star$ & \textbf{47.3}$\downarrow$ & \textbf{47.5}$\downarrow$$^\star$ & 69.7$\downarrow$$^\star$ & 64.6$\downarrow$$^\star$ & 62.2$\downarrow$$^\star$ & 45.1$\downarrow$ & 93.8$\uparrow$ & 67.0$\downarrow$ & 63.5$\downarrow$$^\star$ \\
 &  & NutriVisRAG (Nomic Vision) & 41.5$\downarrow$$^\star$ & 40.1$\downarrow$$^\star$ & 53.8$\downarrow$$^\star$ & 40.8$\uparrow$ & 58.3$\downarrow$ & 51.0$\downarrow$ & 50.9$\downarrow$$^\star$ & 74.3$\downarrow$ & 57.5$\downarrow$$^\star$ & 64.1$\downarrow$$^\star$ & 40.8$\downarrow$$^\star$ & 79.2$\downarrow$$^\star$ & \textbf{61.4}$\downarrow$ & \textbf{60.3}$\downarrow$$^\star$ \\
\midrule
 & GPT-5.4-mini & Vanilla & 53.8 & 45.0 & 75.8 & 32.8 & 46.9 & 51.8 & 57.4 & 93.0 & 76.4 & 89.2 & 38.5 & 96.2 & 74.6 & 75.7 \\
 &  & NutriVisRAG (PE-Core) & 27.5$\downarrow$$^\star$ & 31.9$\downarrow$$^\star$ & 50.7$\downarrow$$^\star$ & 33.2$\uparrow$ & 45.2$\downarrow$ & \textbf{43.0}$\downarrow$ & \textbf{44.5}$\downarrow$$^\star$ & 42.3$\downarrow$$^\star$ & 39.9$\downarrow$$^\star$ & 56.5$\downarrow$$^\star$ & 34.5$\downarrow$ & 50.0$\downarrow$$^\star$ & \textbf{47.0}$\downarrow$ & \textbf{48.7}$\downarrow$$^\star$ \\
 &  & NutriVisRAG (SigLIP2) & 29.1$\downarrow$$^\star$ & 31.9$\downarrow$$^\star$ & 51.8$\downarrow$$^\star$ & 35.9$\uparrow$ & 48.9$\uparrow$ & 45.5$\downarrow$ & 46.5$\downarrow$$^\star$ & 41.2$\downarrow$$^\star$ & 44.2$\downarrow$$^\star$ & 60.8$\downarrow$$^\star$ & 36.7$\downarrow$ & 56.5$\downarrow$$^\star$ & 51.3$\downarrow$ & 52.9$\downarrow$$^\star$ \\
 &  & NutriVisRAG (Nomic Vision) & 31.7$\downarrow$$^\star$ & 33.7$\downarrow$$^\star$ & 52.5$\downarrow$$^\star$ & 33.6$\uparrow$ & 49.6$\uparrow$ & 45.2$\downarrow$ & 46.4$\downarrow$$^\star$ & 55.6$\downarrow$$^\star$ & 43.3$\downarrow$$^\star$ & 58.2$\downarrow$$^\star$ & 35.1$\downarrow$ & 59.0$\downarrow$$^\star$ & 50.8$\downarrow$ & 51.6$\downarrow$$^\star$ \\
\midrule
 & Mistral-Small-3.2 & Vanilla & 39.7 & 56.8 & 105.5 & 45.1 & 63.8 & 71.5 & 79.5 & 55.5 & 38.5 & 77.6 & 38.6 & 63.7 & 60.0 & 63.4 \\
 &  & NutriVisRAG (PE-Core) & 48.6$\uparrow$ & 50.1$\downarrow$ & 75.5$\downarrow$$^\star$ & 55.9$\uparrow$ & 64.8$\uparrow$ & \textbf{65.4}$\downarrow$ & \textbf{67.6}$\downarrow$$^\star$ & 37.8$\downarrow$$^\star$ & 40.3$\uparrow$ & 52.6$\downarrow$$^\star$ & 47.9$\uparrow$$^{\blacktriangledown}$ & 60.5$\downarrow$ & \textbf{53.7}$\downarrow$ & \textbf{52.8}$\downarrow$$^\star$ \\
 &  & NutriVisRAG (SigLIP2) & 65.5$\uparrow$$^{\blacktriangledown}$ & 64.7$\uparrow$ & 71.9$\downarrow$$^\star$ & 84.6$\uparrow$$^{\blacktriangledown}$ & 86.9$\uparrow$$^{\blacktriangledown}$ & 81.1$\uparrow$ & 78.6$\downarrow$ & 33.5$\downarrow$$^\star$ & 42.0$\uparrow$ & 52.2$\downarrow$$^\star$ & 52.6$\uparrow$$^{\blacktriangledown}$ & 68.6$\uparrow$ & 57.8$\downarrow$ & 55.6$\downarrow$$^\star$ \\
 &  & NutriVisRAG (Nomic Vision) & 61.9$\uparrow$ & 66.4$\uparrow$ & 90.2$\downarrow$$^\star$ & 69.0$\uparrow$$^{\blacktriangledown}$ & 84.5$\uparrow$ & 81.2$\uparrow$ & 82.8$\uparrow$ & 44.4$\downarrow$$^\star$ & 53.4$\uparrow$ & 72.6$\downarrow$$^\star$ & 59.3$\uparrow$ & 75.1$\uparrow$ & 69.0$\uparrow$ & 69.2$\uparrow$ \\
\midrule
 & GLM-5.3-Flash & Vanilla & 50.5 & 41.9 & 60.5 & 41.5 & 43.6 & 48.5 & 51.5 & 51.5 & 37.7 & 58.4 & 34.5 & 42.4 & 45.1 & 48.2 \\
 &  & NutriVisRAG (PE-Core) & 32.9$\downarrow$$^\star$ & 31.9$\downarrow$$^\star$ & 42.9$\downarrow$$^\star$ & 36.5$\downarrow$ & 42.3$\downarrow$ & 40.6$\downarrow$ & 40.9$\downarrow$$^\star$ & 33.4$\downarrow$$^\star$ & 31.5$\downarrow$$^\star$ & 42.6$\downarrow$$^\star$ & 34.5$\uparrow$ & 42.7$\uparrow$ & 39.9$\downarrow$ & 40.2$\downarrow$$^\star$ \\
 &  & NutriVisRAG (SigLIP2) & 31.6$\downarrow$$^\star$ & 29.9$\downarrow$$^\star$ & 43.1$\downarrow$$^\star$ & 33.4$\downarrow$$^\star$ & 40.3$\downarrow$ & \textbf{38.9}$\downarrow$ & \textbf{39.7}$\downarrow$$^\star$ & 32.5$\downarrow$$^\star$ & 31.1$\downarrow$$^\star$ & 41.9$\downarrow$$^\star$ & 32.4$\downarrow$ & 41.0$\downarrow$ & \textbf{38.4}$\downarrow$ & \textbf{38.9}$\downarrow$$^\star$ \\
 &  & NutriVisRAG (Nomic Vision) & 35.3$\downarrow$$^\star$ & 33.3$\downarrow$$^\star$ & 49.2$\downarrow$$^\star$ & 31.0$\downarrow$$^\star$ & 43.9$\uparrow$ & 41.4$\downarrow$ & 42.8$\downarrow$$^\star$ & 37.3$\downarrow$$^\star$ & 33.4$\downarrow$$^\star$ & 47.1$\downarrow$$^\star$ & 33.8$\downarrow$ & 42.2$\downarrow$ & 41.1$\downarrow$ & 42.2$\downarrow$$^\star$ \\
\midrule
EDUCATE & Gemma-4-31B & Vanilla & -- & -- & \textbf{33.0} & -- & -- & -- & -- & -- & -- & 45.4 & -- & -- & -- & -- \\
 &  & NutriVisRAG (PE-Core) & -- & -- & 36.5$\uparrow$$^{\blacktriangledown}$ & -- & -- & -- & -- & -- & -- & 45.0$\downarrow$ & -- & -- & -- & -- \\
 &  & NutriVisRAG (SigLIP2) & -- & -- & 37.8$\uparrow$ & -- & -- & -- & -- & -- & -- & \textbf{44.5}$\downarrow$ & -- & -- & -- & -- \\
 &  & NutriVisRAG (Nomic Vision) & -- & -- & 38.5$\uparrow$ & -- & -- & -- & -- & -- & -- & 47.0$\uparrow$ & -- & -- & -- & -- \\
\midrule
 & GPT-5.4-mini & Vanilla & -- & -- & 45.0 & -- & -- & -- & -- & -- & -- & 58.4 & -- & -- & -- & -- \\
 &  & NutriVisRAG (PE-Core) & -- & -- & 40.9$\downarrow$ & -- & -- & -- & -- & -- & -- & 46.6$\downarrow$$^\star$ & -- & -- & -- & -- \\
 &  & NutriVisRAG (SigLIP2) & -- & -- & \textbf{38.0}$\downarrow$$^\star$ & -- & -- & -- & -- & -- & -- & \textbf{43.3}$\downarrow$$^\star$ & -- & -- & -- & -- \\
 &  & NutriVisRAG (Nomic Vision) & -- & -- & 42.5$\downarrow$ & -- & -- & -- & -- & -- & -- & 43.6$\downarrow$$^\star$ & -- & -- & -- & -- \\
\midrule
 & Mistral-Small-3.2 & Vanilla & -- & -- & \textbf{51.2} & -- & -- & -- & -- & -- & -- & \textbf{51.9} & -- & -- & -- & -- \\
 &  & NutriVisRAG (PE-Core) & -- & -- & 153.8$\uparrow$$^{\blacktriangledown}$ & -- & -- & -- & -- & -- & -- & 82.2$\uparrow$$^{\blacktriangledown}$ & -- & -- & -- & -- \\
 &  & NutriVisRAG (SigLIP2) & -- & -- & 151.6$\uparrow$$^{\blacktriangledown}$ & -- & -- & -- & -- & -- & -- & 92.8$\uparrow$$^{\blacktriangledown}$ & -- & -- & -- & -- \\
 &  & NutriVisRAG (Nomic Vision) & -- & -- & 147.1$\uparrow$$^{\blacktriangledown}$ & -- & -- & -- & -- & -- & -- & 81.4$\uparrow$$^{\blacktriangledown}$ & -- & -- & -- & -- \\
\midrule
 & GLM-5.3-Flash & Vanilla & -- & -- & 53.6 & -- & -- & -- & -- & -- & -- & 54.0 & -- & -- & -- & -- \\
 &  & NutriVisRAG (PE-Core) & -- & -- & \textbf{47.8}$\downarrow$$^\star$ & -- & -- & -- & -- & -- & -- & 44.0$\downarrow$$^\star$ & -- & -- & -- & -- \\
 &  & NutriVisRAG (SigLIP2) & -- & -- & 47.9$\downarrow$$^\star$ & -- & -- & -- & -- & -- & -- & 44.1$\downarrow$$^\star$ & -- & -- & -- & -- \\
 &  & NutriVisRAG (Nomic Vision) & -- & -- & 53.2$\downarrow$ & -- & -- & -- & -- & -- & -- & \textbf{42.0}$\downarrow$$^\star$ & -- & -- & -- & -- \\
\bottomrule
\end{tabular}%
}
\end{table}

Furthermore, to provide a compact cross-dataset summary, Table~\ref{tab:cross-dataset-average} reports the unweighted mean of the dataset-level $\mathrm{\%MAE}_{\mathrm{micro}}$ values across Nutrition5k, SNAPMe, and MADiMa.
An unweighted mean is used so that each benchmark contributes equally rather than allowing the larger Nutrition5k query set to dominate the summary.
These values are descriptive and should not be interpreted as pooled micro errors or as evidence that a single retrieval encoder is uniformly best across query sets.
EDUCATE is excluded because only CHO is available, whereas the cross-dataset summary is based on $\mathrm{\%MAE}_{\mathrm{micro}}$.
Across the eight VLM--prompt combinations, PE-Core has the lowest descriptive mean in four, SigLIP\,2 in three, and Vanilla in one.
These aggregate differences do not establish consistent superiority across individual query sets.

\begin{table}[tb]
\centering
\caption{Descriptive unweighted cross-dataset mean $\mathrm{\%MAE}_{\mathrm{micro}}$ for each VLM, prompting strategy, and inference setting across Nutrition5k, SNAPMe, and MADiMa (lower is better).
}
\label{tab:cross-dataset-average}
\small
\setlength{\tabcolsep}{4pt}
\resizebox{0.75\textwidth}{!}{%
\begin{tabular}{llrrrr}
\toprule
\textbf{VLM} & \textbf{Prompt} & \textbf{Vanilla} & \textbf{PE-Core} & \textbf{SigLIP\,2} & \textbf{Nomic Vision} \\
\midrule
Gemma-4-31B & Per-item & 49.8 & 50.1 & \textbf{48.7} & 51.9 \\
 & Whole-meal & 59.3 & 52.1 & \textbf{51.6} & 52.6 \\
\midrule
GPT-5.4-mini & Per-item & 52.2 & \textbf{43.0} & 44.0 & 45.4 \\
 & Whole-meal & 58.5 & \textbf{45.0} & 46.0 & 47.5 \\
\midrule
Mistral-Small-3.2 & Per-item & \textbf{67.5} & 94.1 & 109.5 & 112.2 \\
 & Whole-meal & 62.2 & \textbf{60.8} & 64.7 & 69.9 \\
\midrule
GLM-5.3-Flash & Per-item & 48.4 & \textbf{45.0} & 45.6 & 47.6 \\
 & Whole-meal & 48.0 & 45.7 & \textbf{44.7} & 47.7 \\
\bottomrule
\end{tabular}
}
\end{table}

\section{Retrieval Diagnostics}
\label{sec:retrieval-diagnostics}

Unless otherwise stated, the exemplar-level diagnostics use the main retrieval configuration with $K=5$, $\tau=0.75$, and the closest-reference fallback.
Retrieval benefit is defined as the per-query reduction in pooled absolute macronutrient error relative to Vanilla, normalised by the query-set mean target macronutrient sum; EDUCATE instead uses absolute CHO error normalised by mean annotated CHO.
Visual similarity is the mean cosine similarity between the query and its retrieved exemplars.
Nutritional distance is the mean absolute query--exemplar difference across the available macronutrients after normalising each nutrient by its query-set mean; EDUCATE uses CHO only.
Associations are quantified using Spearman's rank coefficient ($\rho$).

\subsection{Visual Retrieval Space}
\label{subsec:embedding-space}

\begin{figure}[tb]
\centering
\includegraphics[width=0.9\textwidth]{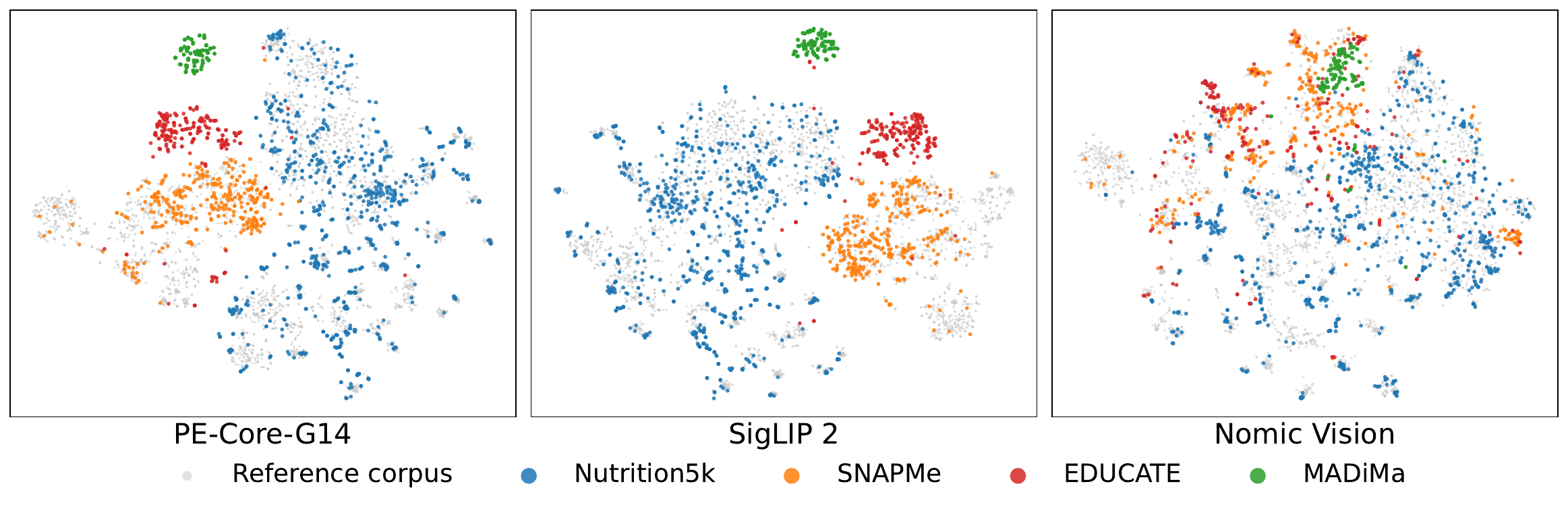}
\caption{t-SNE visualisation of the retrieval embedding spaces for PE-Core-G14, SigLIP\,2, and Nomic Vision.
Gray points represent the combined Nutrition5k--SNAPMe reference corpus, while coloured points represent evaluation queries from Nutrition5k, SNAPMe, EDUCATE, and MADiMa.}
\label{fig:embed}
\end{figure}

Figure~\ref{fig:embed} qualitatively examines whether projected query--reference overlap in the visual retrieval embedding space corresponds to downstream NutriVisRAG gains.
For each encoder, t-SNE (perplexity 30, PCA initialisation, random seed 0) is applied with cosine distance jointly to the reference corpus and all query sets.
PE-Core and SigLIP\,2 show stronger dataset-level separation, particularly for MADiMa and EDUCATE, whereas Nomic Vision shows greater mixing between queries and references.
However, projected overlap does not correspond directly to retrieval benefit: MADiMa improves in 22/24 configurations despite its separation under PE-Core and SigLIP\,2, while SNAPMe improves in only 7/24 despite its training split contributing to the reference corpus.
Likewise, Nomic Vision's greater mixing does not yield more consistent MADiMa gains, improving 6/8 configurations compared with 8/8 for PE-Core and SigLIP\,2.
Because t-SNE can distort neighbourhood and global structure, this analysis is interpreted qualitatively and motivates the exemplar-level analyses below.

\subsection{VLM-Specific Similarity--Benefit Associations}
\label{subsec:sup-similarity-benefit}

\begin{figure}[tb]
\centering
\includegraphics[width=0.82\textwidth]{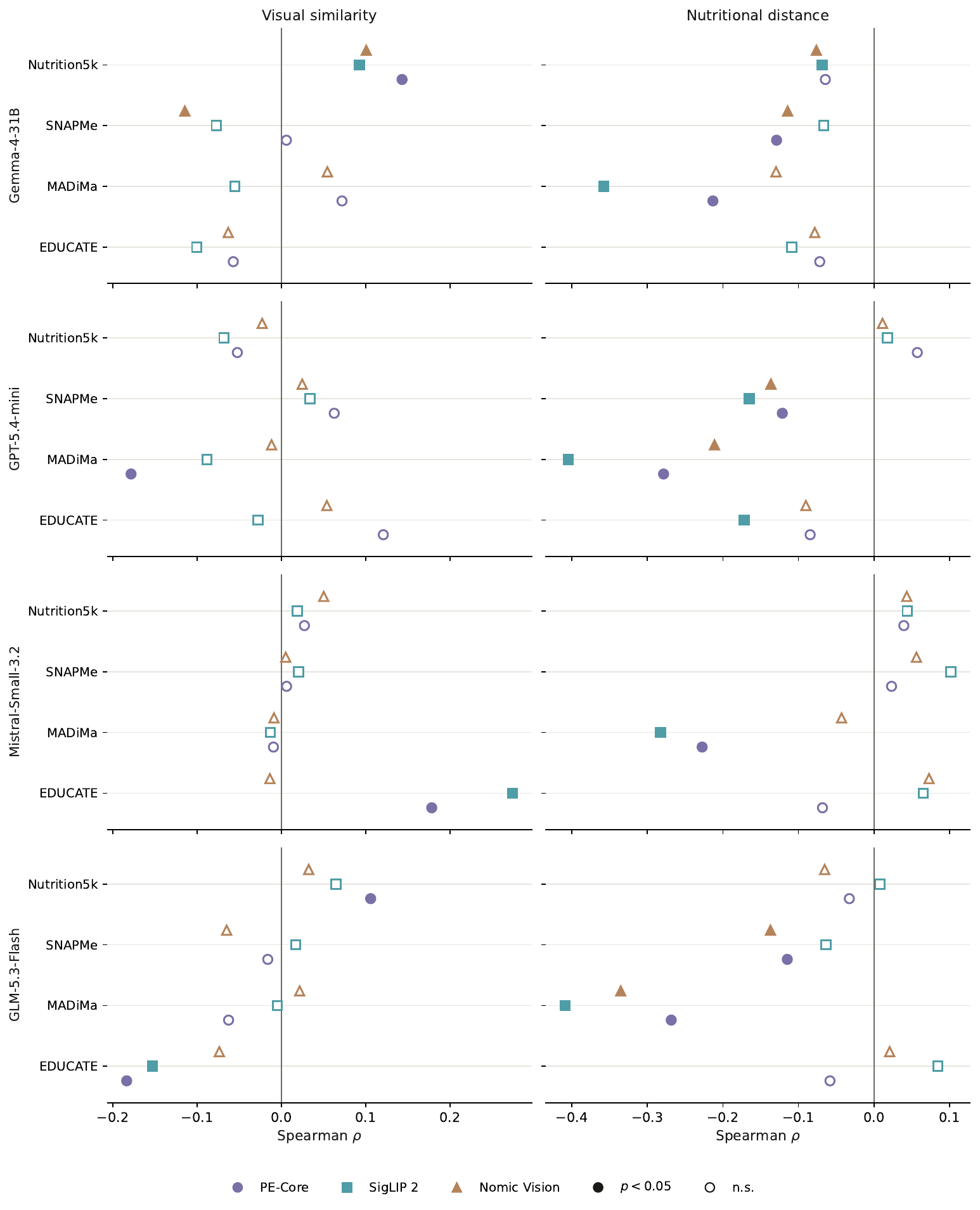}
\caption{VLM-specific Spearman associations between retrieval benefit and mean visual similarity (left) or retrospective nutritional distance (right) under per-item prompting.
Rows represent VLMs, and markers represent retrieval encoders.
Filled markers indicate Benjamini--Hochberg-corrected $q<0.05$; open markers indicate non-significant associations.}
\label{fig:sup-rho-peritem}
\end{figure}

\begin{figure}[tb]
\centering
\includegraphics[width=0.82\textwidth]{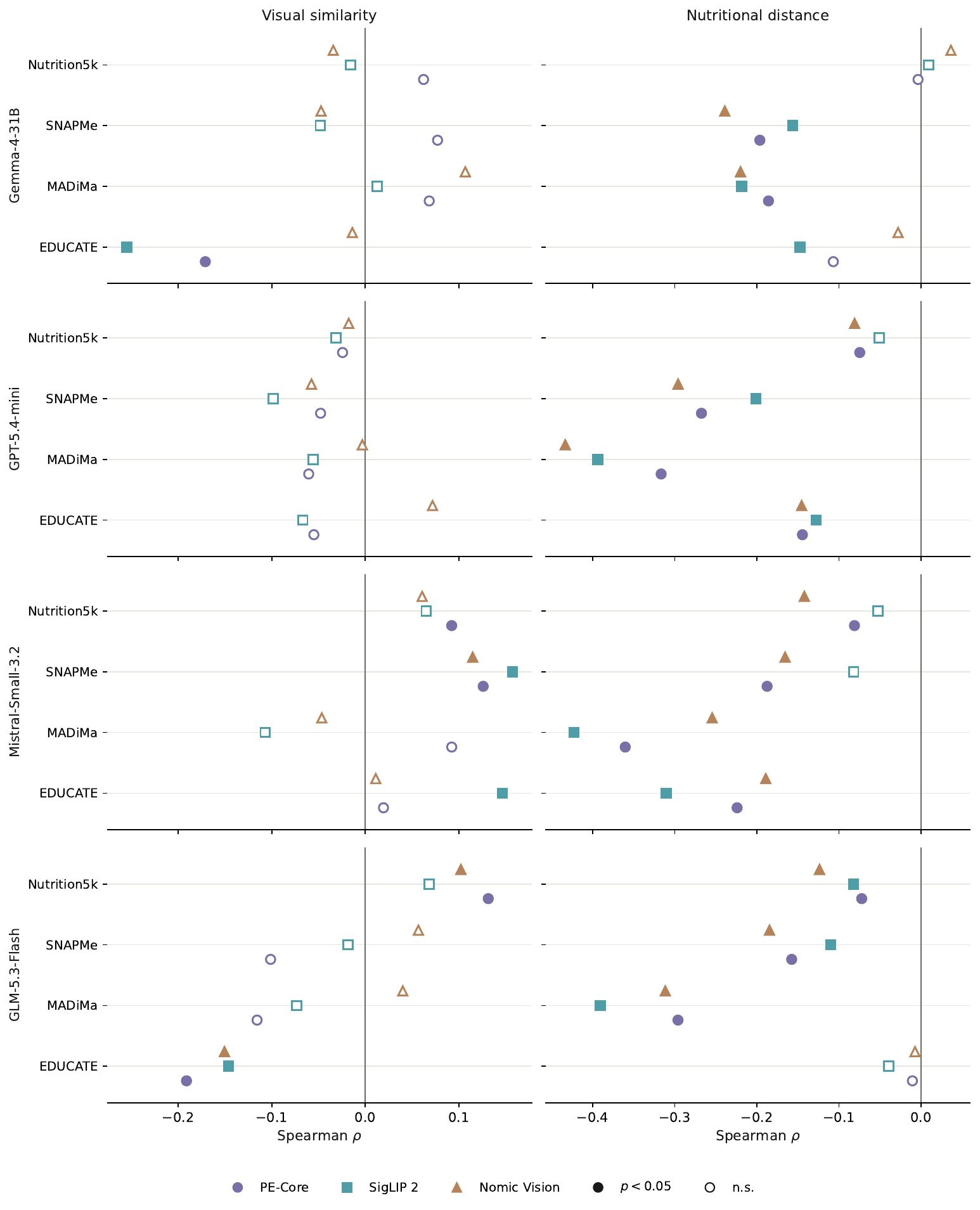}
\caption{VLM-specific Spearman associations between retrieval benefit and mean visual similarity (left) or retrospective nutritional distance (right) under whole-meal prompting.
Rows represent VLMs, and markers represent retrieval encoders.
Filled markers indicate Benjamini--Hochberg-corrected $q<0.05$; open markers indicate non-significant associations.}
\label{fig:sup-rho-wholemeal}
\end{figure}

The main paper summarises the association between retrieval benefit and exemplar similarity after averaging retrieval benefit across the four VLMs.
Figures~\ref{fig:sup-rho-peritem} and~\ref{fig:sup-rho-wholemeal} show the corresponding associations separately for each VLM under per-item and whole-meal prompting.

Visual similarity remains weak and inconsistent across VLMs, query sets, and retrieval encoders under both prompting strategies.
Nutritional distance shows more consistently negative associations under whole-meal prompting, particularly on MADiMa, whereas the relationship is weaker and less consistent under per-item prompting.
The VLM-specific results further show that retrieval benefit varies across models despite using the same retrieved evidence, consistent with the configuration-dependent effects observed in the main experiments.

\subsection{Visual Similarity and Nutritional Proximity}
\label{subsec:visual-nutritional-proximity}

\begin{figure}[tb]
\centering
\includegraphics[width=0.75\textwidth]{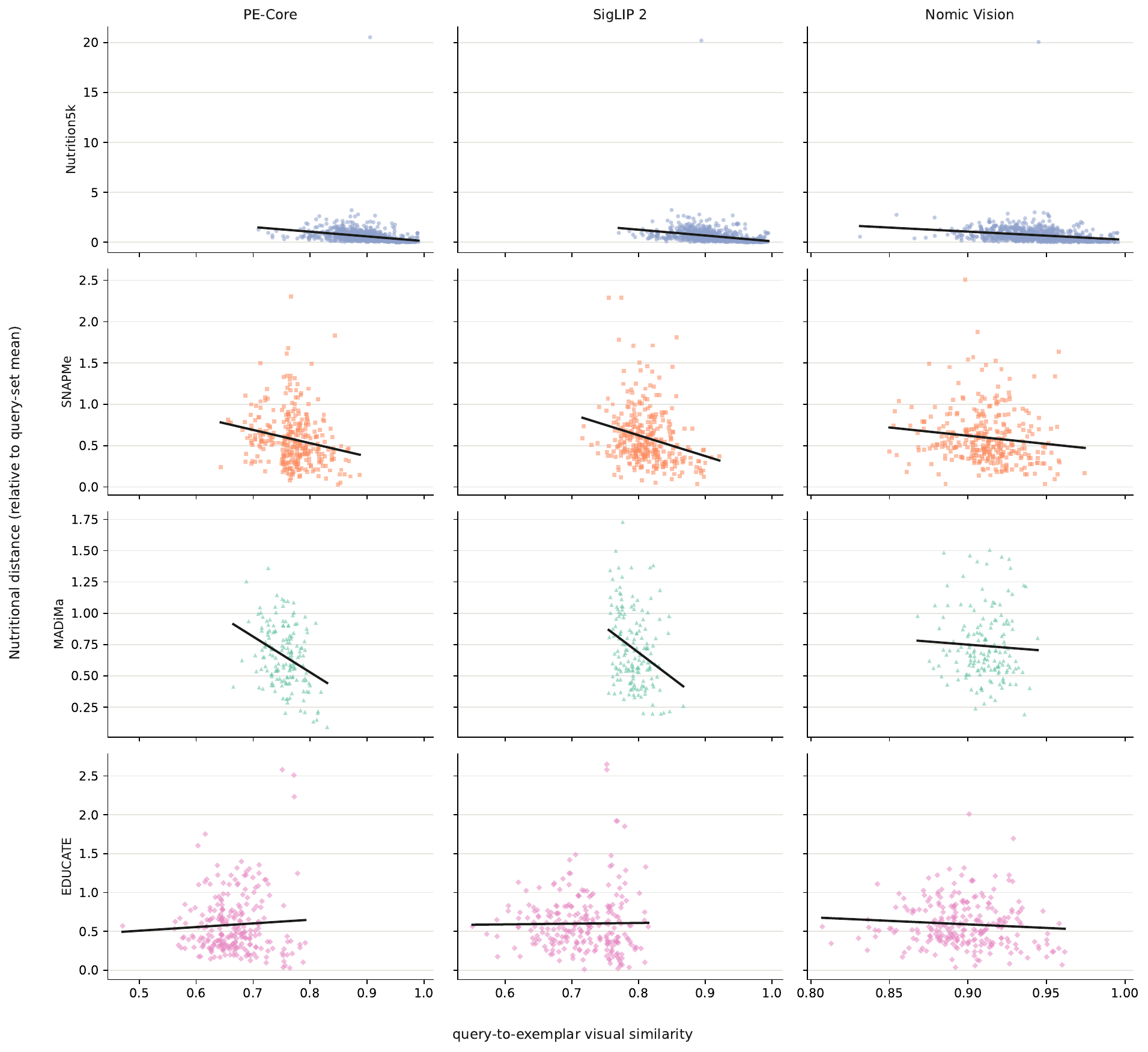}
\caption{Mean visual similarity versus normalised nutritional distance between queries and retrieved exemplars.
Rows represent query sets and columns represent PE-Core, SigLIP\,2, and Nomic Vision.
Lower nutritional distance indicates greater nutritional proximity.}
\label{fig:sup-sim-distance}
\end{figure}

Because each retrieved meal image is provided together with its nutritional annotations, we additionally examine whether visually closer exemplars are also nutritionally closer to the query.
This analysis is independent of prompting strategy because the retrieved exemplars are fixed before VLM prompting.
Figure~\ref{fig:sup-sim-distance} shows that the relationship varies substantially across query sets.
Visual similarity is more strongly associated with nutritional proximity on Nutrition5k, with weaker associations on SNAPMe and MADiMa and little association on EDUCATE.
Thus, visually similar retrieval does not uniformly imply quantitatively similar nutritional annotations across datasets.
This provides additional context for the weak and inconsistent relationship between cosine similarity and downstream retrieval benefit.

Nutritional distance is calculated retrospectively from annotated query nutrients and is therefore unavailable as an inference-time retrieval criterion.
Moreover, nutritional distance and estimation error both depend on the annotated query targets, so their association with retrieval benefit should be interpreted descriptively rather than causally.

\subsection{Meal-Level Error and Agreement Analysis}
\label{sec:meal-level-analysis}

Having characterised the retrieved evidence and its association with estimation benefit, we next examine how these effects are distributed across individual meals.
We analyse NutriVisRAG at the meal level under both per-item and whole-meal prompting.
Whole-meal prompting shows more consistent retrieval benefits in the main experiments, reducing the primary error in 35 of 48 configurations compared with 21 of 48 under per-item prompting.
To avoid selecting a particular retrieval representation or VLM, the analysis includes all three visual encoders, all four VLMs, and all four query sets.
For the paired-error analysis, the per-meal quantity underlying $\mathrm{\%MAE}_{\text{micro}}$ is the pooled absolute macronutrient error, $|\Delta\mathrm{CHO}|+|\Delta\mathrm{PROT}|+|\Delta\mathrm{FAT}|$, expressed in grams; for EDUCATE, the corresponding quantity is absolute CHO error.
For the Bland--Altman analysis, the corresponding pooled predicted and reference macronutrient quantities, $\widehat{\mathrm{CHO}}+\widehat{\mathrm{PROT}}+\widehat{\mathrm{FAT}}$ and $\mathrm{CHO}+\mathrm{PROT}+\mathrm{FAT}$, are compared in grams, while EDUCATE uses predicted and reference CHO.
The analysis uses the main retrieval configuration throughout: the combined Nutrition5k--SNAPMe reference corpus, $K=5$, and the similarity threshold $\tau=0.75$ with the closest-reference fallback where applicable.

\begin{sidewaysfigure}[tb]
\centering
\includegraphics[width=\textwidth]{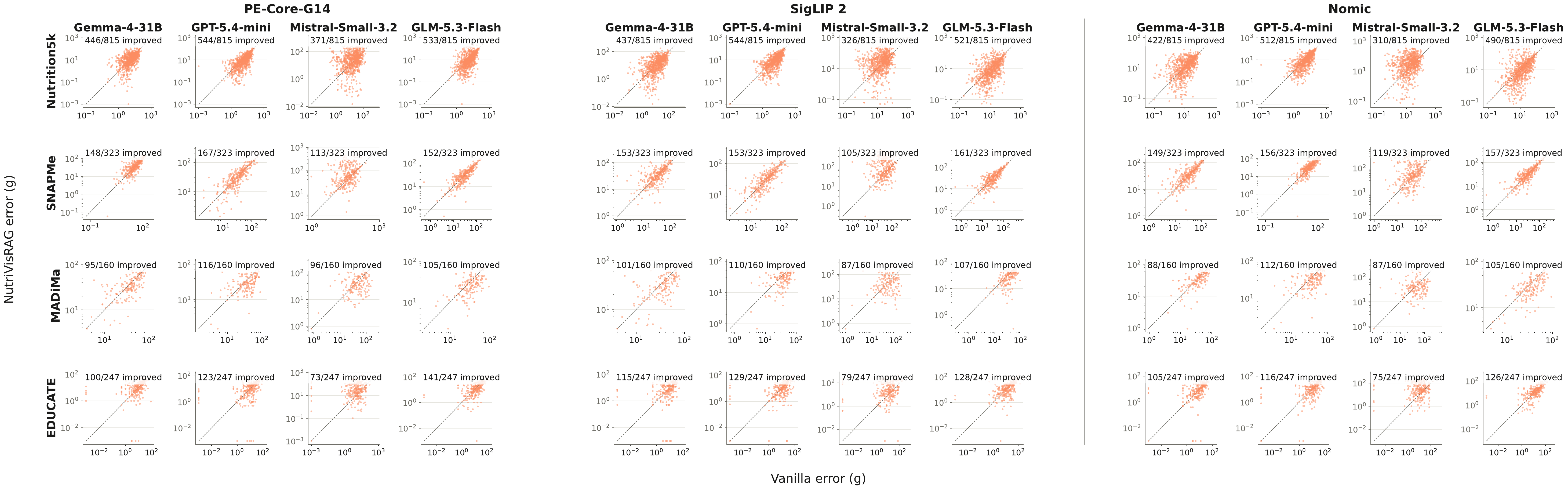}
\caption{
Meal-level absolute error of Vanilla versus NutriVisRAG under per-item prompting.
Error is $|\Delta\mathrm{CHO}|+|\Delta\mathrm{PROT}|+|\Delta\mathrm{FAT}|$ in grams (absolute CHO error for EDUCATE).
Rows: query sets; columns: retrieval encoders, each with Gemma-4-31B, GPT-5.4-mini, Mistral-Small-3.2, and GLM-5.3-Flash.
Points below the diagonal favour NutriVisRAG, and each panel reports the number of improved meals.
Both axes are logarithmic, with $10^{-3}$\,g added to all errors so that zero errors remain visible.
}
\label{fig:paired-errors-peritem}
\end{sidewaysfigure}

\begin{sidewaysfigure}[tb]
\centering
\includegraphics[width=\textwidth]{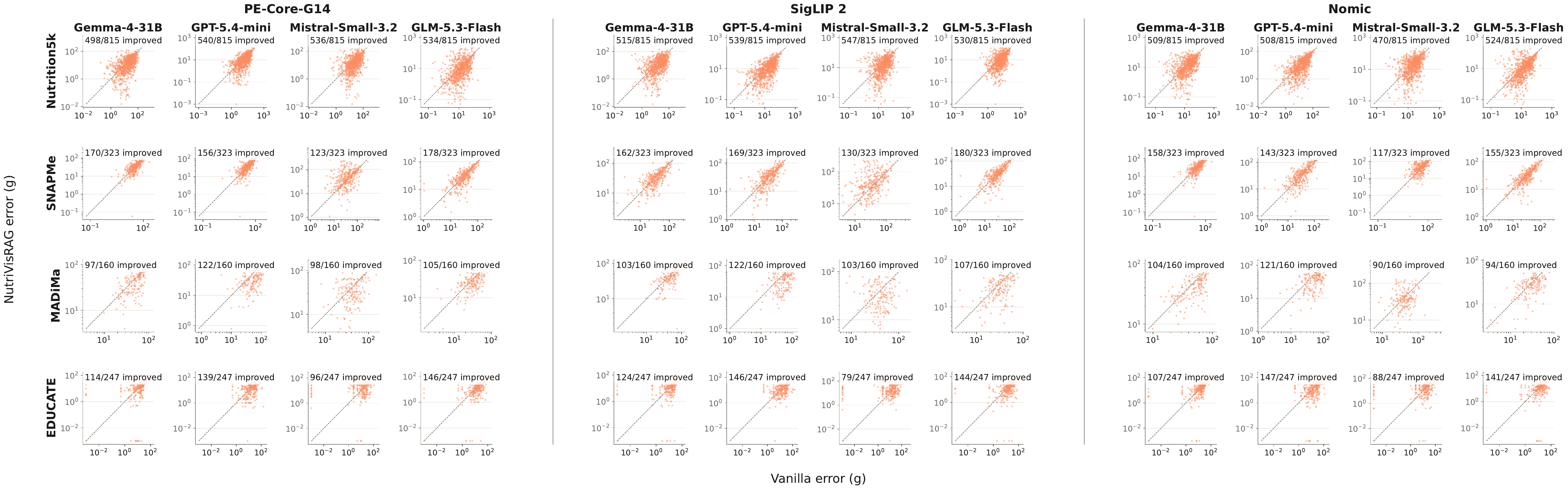}
\caption{
Meal-level absolute error of Vanilla versus NutriVisRAG under whole-meal prompting.
Error is $|\Delta\mathrm{CHO}|+|\Delta\mathrm{PROT}|+|\Delta\mathrm{FAT}|$ in grams (absolute CHO error for EDUCATE).
Rows: query sets; columns: retrieval encoders, each with Gemma-4-31B, GPT-5.4-mini, Mistral-Small-3.2, and GLM-5.3-Flash.
Points below the diagonal favour NutriVisRAG, and each panel reports the number of improved meals.
Both axes are logarithmic, with $10^{-3}$\,g added to all errors so that zero errors remain visible.
}
\label{fig:paired-errors}
\end{sidewaysfigure}

Figures~\ref{fig:paired-errors-peritem} and~\ref{fig:paired-errors} compare the absolute estimation error of Vanilla and NutriVisRAG for each meal under per-item and whole-meal prompting, respectively.
Whole-meal prompting yields a larger number of individually improved meals than per-item prompting in 37 of the 48 matched VLM--encoder--query-set panels.
More than half of the meals improve in 36/48 whole-meal panels compared with 26/48 per-item panels, consistent with the greater aggregate benefit of whole-meal retrieval.
However, the number of improved meals does not necessarily determine the aggregate error because $\mathrm{\%MAE}_{\text{micro/CHO}}$ also depends on the magnitude of individual errors.
For example, GLM-5.3-Flash under whole-meal prompting improves more than half of Nutrition5k meals for all three retrieval encoders despite increasing the aggregate primary error in all three configurations.
The paired plots therefore show that retrieval can benefit a majority of meals while a smaller number of large degradations dominate the aggregate error.

\begin{sidewaysfigure}[tb]
\centering
\includegraphics[width=\textwidth]{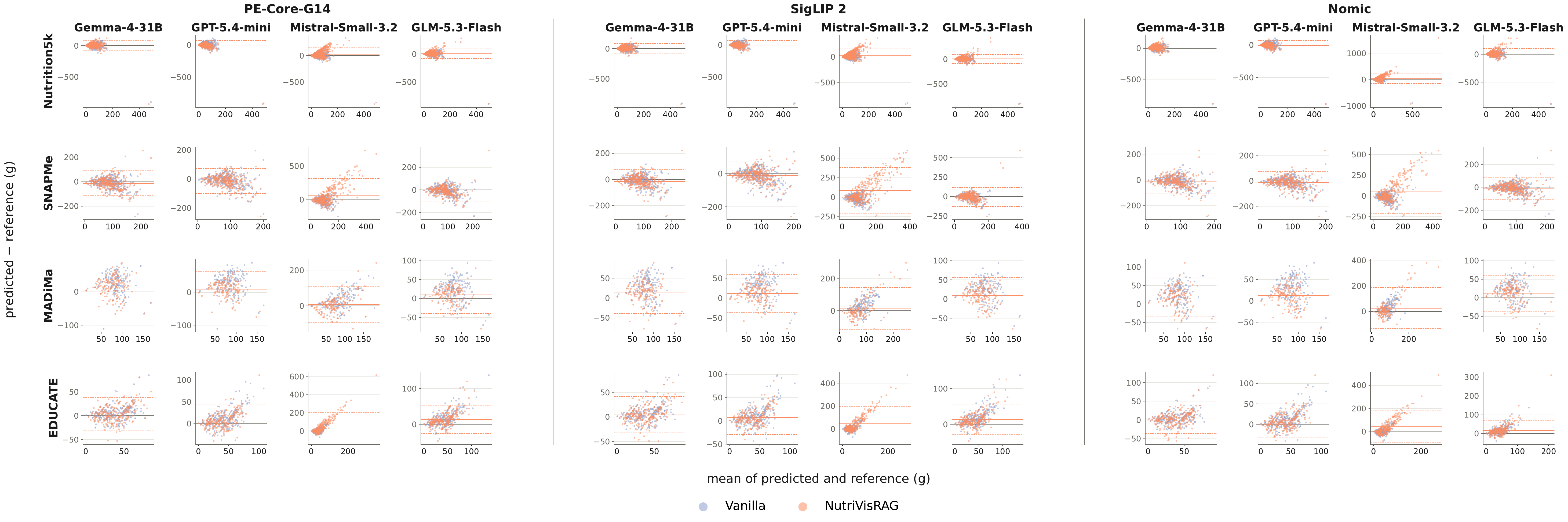}
\caption{
Bland--Altman analysis of Vanilla and NutriVisRAG predictions under per-item prompting.
The analysed quantity is $\mathrm{CHO}+\mathrm{PROT}+\mathrm{FAT}$ in grams (CHO for EDUCATE); panel layout as in Fig.~\ref{fig:paired-errors-peritem}.
Each point shows the difference (predicted minus reference) against the mean of the two.
For each condition, solid and dashed lines show the mean difference and $\pm1.96$\,SD limits of agreement, respectively.
}
\label{fig:bland-altman-peritem}
\end{sidewaysfigure}

\begin{sidewaysfigure}[tb]
\centering
\includegraphics[width=\textwidth]{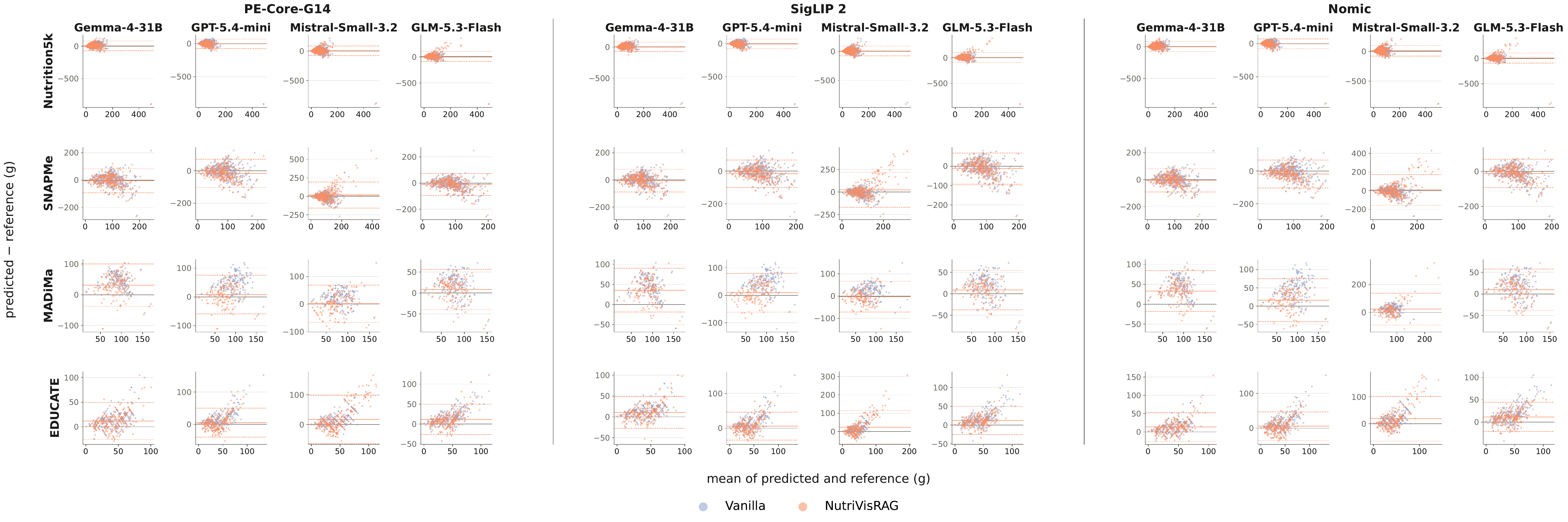}
\caption{
Bland--Altman analysis of Vanilla and NutriVisRAG predictions under whole-meal prompting.
The analysed quantity is $\mathrm{CHO}+\mathrm{PROT}+\mathrm{FAT}$ in grams (CHO for EDUCATE); panel layout as in Fig.~\ref{fig:paired-errors}.
Each point shows the difference (predicted minus reference) against the mean of the two.
For each condition, solid and dashed lines show the mean difference and $\pm1.96$\,SD limits of agreement, respectively.
}
\label{fig:bland-altman}
\end{sidewaysfigure}

Figures~\ref{fig:bland-altman-peritem} and~\ref{fig:bland-altman} provide complementary Bland--Altman analyses of the corresponding predictions under per-item and whole-meal prompting, respectively.
The plots show that retrieval affects not only average error but also systematic bias and dispersion, with several configurations exhibiting increasing disagreement at larger meal magnitudes.
This magnitude-dependent behaviour is particularly pronounced for Mistral-Small-3.2 on SNAPMe and EDUCATE, where large positive deviations remain visible across retrieval encoders.
Substantial between-meal dispersion and isolated large errors are also present in configurations with lower aggregate error, showing that improved average performance does not imply uniformly improved agreement.
The limits of agreement are therefore interpreted descriptively because the differences are right-skewed in several configurations and can be strongly influenced by large-error meals.
For SNAPMe, the reference nutrients are verified self-reported dietary estimates, whereas EDUCATE uses the mean CHO estimate of two dietitians; disagreement on these datasets therefore also incorporates uncertainty in the reference annotations.
Together, the paired-error and agreement analyses show that the effect of visual exemplar grounding depends on both how many meals improve and how strongly individual meal errors change.


\section{Exploratory Analyses}
\label{sec:exploratory}

In addition to the main retrieval experiments and ablations, we investigate two exploratory extensions across all four evaluated VLMs.
Post-hoc food-composition grounding replaces VLM-predicted nutrient values with database-derived estimates based on predicted food identities and weights, and is therefore evaluated under per-item prompting, which provides these predictions.
Retrieval reranking instead modifies exemplar selection by rescoring visually retrieved candidates and is evaluated under both prompting strategies.

\subsection{Post-Hoc Food-Composition Grounding}
\label{sec:qepost}

We investigate whether nutrient estimation can be improved by replacing the VLM-predicted nutrient values with values derived from a food-composition database.
This procedure requires individually identified foods and their estimated quantities, so the analysis is restricted to per-item predictions.
Each VLM-predicted food name and weight are retained; the food name is matched to a food-composition entry and scaled by the predicted weight, and the resulting item-level values are summed to obtain meal-level estimates.

\paragraph{Food-name matching.}
We compare one lexical and three embedding-based matching strategies.
Lexical matching selects the database entry with the highest token-set similarity to the VLM-predicted food name.
The embedding-based strategies encode the predicted food name and database entry names and select the entry with the highest cosine similarity.
We evaluate Qwen3-Embedding~\cite{zhang2025qwen3}, EmbeddingGemma~\cite{vera2025embeddinggemma} and Nomic Embed Text v2 (MoE) (Nomic Text)~\cite{nussbaum2025training}.
For all strategies, the highest-scoring database entry is accepted without an additional threshold.

\paragraph{Food-composition databases.}
For Nutrition5k, SNAPMe, and MADiMa, nutrient values are obtained from the U.S. Department of Agriculture (USDA) FoodData Central (FDC)~\cite{fukagawa2022fooddata}.
For EDUCATE, the predicted food name is first matched against the Dutch Food Composition Database (NEVO)~\cite{nevo2025}, with FDC used as a fallback when the highest NEVO matching score is below 0.5.
This NEVO-based configuration yields lower CHO error than FDC-only lookup across the evaluated EDUCATE configurations and is therefore reported in Table~\ref{tab:post-grounding}.

\paragraph{Experimental setup.}
The analysis uses the same fixed 100-query subsets as the retrieval ablations, with $K=5$ and $\tau=0.75$.
The same source prediction is used for the original NutriVisRAG estimate and all post-hoc variants, so only the food-name matching and nutrient-substitution step changes between conditions.
Embedding models are run locally on an NVIDIA RTX A6000.

\begin{table}[tb]
\centering
\caption{Post-hoc food-composition grounding under per-item prompting on the fixed 100-query subsets (\%MAE, lower is better).
NutriVisRAG denotes the original prediction; the other columns are the matching approaches defined in the text.
Dashes indicate unavailable evaluation targets for the corresponding dataset.
Bold indicates the lowest error per row and metric; $^\star$/$^{\blacktriangledown}$ indicate significant improvements/degradations relative to NutriVisRAG (paired Wilcoxon signed-rank test, Benjamini--Hochberg corrected, $q<0.05$).}
\label{tab:post-grounding}
\scriptsize
\setlength{\tabcolsep}{1.25pt}
\renewcommand{\arraystretch}{0.94}
\resizebox{\textwidth}{!}{%
\begin{tabular}{lllrrrrrrrrrr}
\toprule
& & & \multicolumn{5}{c}{\textbf{Energy}} & \multicolumn{5}{c}{\textbf{Micro / CHO}} \\
\cmidrule(lr){4-8}\cmidrule(lr){9-13}
\textbf{Query set} & \textbf{Model} & \textbf{Encoder} & \textbf{NutriVisRAG} & \textbf{Lexical} & \textbf{Qwen3} & \textbf{Gemma} & \textbf{Nomic Text} & \textbf{NutriVisRAG} & \textbf{Lexical} & \textbf{Qwen3} & \textbf{Gemma} & \textbf{Nomic Text} \\
\midrule
Nutrition5k & GPT-5.4-mini & PE-Core & \textbf{32.8} & 96.0$^{\blacktriangledown}$ & 57.9 & 90.6$^{\blacktriangledown}$ & 78.3$^{\blacktriangledown}$ & \textbf{40.0} & 131.9$^{\blacktriangledown}$ & 61.5 & 132.9$^{\blacktriangledown}$ & 83.4$^{\blacktriangledown}$ \\
 &  & SigLIP2 & \textbf{30.2} & 147.3$^{\blacktriangledown}$ & 48.2 & 82.7$^{\blacktriangledown}$ & 77.7$^{\blacktriangledown}$ & \textbf{35.6} & 172.7$^{\blacktriangledown}$ & 54.3 & 126.8$^{\blacktriangledown}$ & 86.0$^{\blacktriangledown}$ \\
 &  & Nomic Vision & \textbf{29.9} & 127.7$^{\blacktriangledown}$ & 54.5$^{\blacktriangledown}$ & 90.4$^{\blacktriangledown}$ & 80.8$^{\blacktriangledown}$ & \textbf{37.6} & 161.2$^{\blacktriangledown}$ & 57.6$^{\blacktriangledown}$ & 137.4$^{\blacktriangledown}$ & 81.5$^{\blacktriangledown}$ \\
\cmidrule(lr){2-13}
 & Gemma-4-31B & PE-Core & \textbf{50.1} & 144.4$^{\blacktriangledown}$ & 70.1 & 117.9$^{\blacktriangledown}$ & 82.9$^{\blacktriangledown}$ & \textbf{60.4} & 176.2$^{\blacktriangledown}$ & 84.5 & 164.5$^{\blacktriangledown}$ & 96.4$^{\blacktriangledown}$ \\
 &  & SigLIP2 & \textbf{44.8} & 128.7$^{\blacktriangledown}$ & 59.0 & 80.5$^{\blacktriangledown}$ & 74.2$^{\blacktriangledown}$ & \textbf{49.8} & 147.6$^{\blacktriangledown}$ & 64.0 & 106.5$^{\blacktriangledown}$ & 82.2$^{\blacktriangledown}$ \\
 &  & Nomic Vision & \textbf{72.2} & 185.6$^{\blacktriangledown}$ & 78.6$^{\blacktriangledown}$ & 126.0$^{\blacktriangledown}$ & 134.1$^{\blacktriangledown}$ & \textbf{80.7} & 206.8$^{\blacktriangledown}$ & 91.1 & 164.3$^{\blacktriangledown}$ & 124.4$^{\blacktriangledown}$ \\
\cmidrule(lr){2-13}
 & Mistral-Small-3.2 & PE-Core & \textbf{244.8} & 384.4$^{\blacktriangledown}$ & 249.5 & 394.1$^{\blacktriangledown}$ & 311.9 & \textbf{259.9} & 418.2$^{\blacktriangledown}$ & 266.9 & 482.6$^{\blacktriangledown}$ & 346.3 \\
 &  & SigLIP2 & 283.1 & 401.7$^{\blacktriangledown}$ & \textbf{269.6} & 514.0$^{\blacktriangledown}$ & 386.5$^{\blacktriangledown}$ & 306.5 & 434.9$^{\blacktriangledown}$ & \textbf{304.0} & 641.2$^{\blacktriangledown}$ & 445.4 \\
 &  & Nomic Vision & 418.6 & 541.8$^{\blacktriangledown}$ & \textbf{400.1} & 600.7$^{\blacktriangledown}$ & 489.3$^{\blacktriangledown}$ & 410.9 & 524.7$^{\blacktriangledown}$ & \textbf{405.5} & 715.1$^{\blacktriangledown}$ & 491.8 \\
\cmidrule(lr){2-13}
 & GLM-5.3-Flash & PE-Core & \textbf{50.8} & 148.7$^{\blacktriangledown}$ & 88.2$^{\blacktriangledown}$ & 110.5$^{\blacktriangledown}$ & 90.1$^{\blacktriangledown}$ & \textbf{52.5} & 181.8$^{\blacktriangledown}$ & 81.6$^{\blacktriangledown}$ & 156.2$^{\blacktriangledown}$ & 90.7$^{\blacktriangledown}$ \\
 &  & SigLIP2 & \textbf{36.5} & 112.4$^{\blacktriangledown}$ & 59.1$^{\blacktriangledown}$ & 108.6$^{\blacktriangledown}$ & 61.2$^{\blacktriangledown}$ & \textbf{40.7} & 147.9$^{\blacktriangledown}$ & 61.3$^{\blacktriangledown}$ & 161.6$^{\blacktriangledown}$ & 67.7$^{\blacktriangledown}$ \\
 &  & Nomic Vision & \textbf{36.3} & 101.6$^{\blacktriangledown}$ & 42.2$^{\blacktriangledown}$ & 99.5$^{\blacktriangledown}$ & 62.3$^{\blacktriangledown}$ & \textbf{42.0} & 134.3$^{\blacktriangledown}$ & 56.8$^{\blacktriangledown}$ & 139.1$^{\blacktriangledown}$ & 69.4$^{\blacktriangledown}$ \\
\midrule
SNAPMe & GPT-5.4-mini & PE-Core & \textbf{34.5} & 84.0$^{\blacktriangledown}$ & 71.2$^{\blacktriangledown}$ & 58.3$^{\blacktriangledown}$ & 69.3$^{\blacktriangledown}$ & \textbf{42.0} & 109.4$^{\blacktriangledown}$ & 82.3$^{\blacktriangledown}$ & 78.7$^{\blacktriangledown}$ & 78.6$^{\blacktriangledown}$ \\
 &  & SigLIP2 & \textbf{34.5} & 100.6$^{\blacktriangledown}$ & 69.0$^{\blacktriangledown}$ & 64.3$^{\blacktriangledown}$ & 74.8$^{\blacktriangledown}$ & \textbf{42.4} & 119.0$^{\blacktriangledown}$ & 85.0$^{\blacktriangledown}$ & 81.7$^{\blacktriangledown}$ & 84.6$^{\blacktriangledown}$ \\
 &  & Nomic Vision & \textbf{36.1} & 80.6$^{\blacktriangledown}$ & 74.5$^{\blacktriangledown}$ & 61.1$^{\blacktriangledown}$ & 73.7$^{\blacktriangledown}$ & \textbf{42.4} & 105.1$^{\blacktriangledown}$ & 85.5$^{\blacktriangledown}$ & 82.6$^{\blacktriangledown}$ & 84.8$^{\blacktriangledown}$ \\
\cmidrule(lr){2-13}
 & Gemma-4-31B & PE-Core & \textbf{38.8} & 96.5$^{\blacktriangledown}$ & 72.3$^{\blacktriangledown}$ & 54.6$^{\blacktriangledown}$ & 67.6$^{\blacktriangledown}$ & \textbf{44.2} & 112.8$^{\blacktriangledown}$ & 86.6$^{\blacktriangledown}$ & 75.7$^{\blacktriangledown}$ & 76.0$^{\blacktriangledown}$ \\
 &  & SigLIP2 & \textbf{37.9} & 86.0$^{\blacktriangledown}$ & 65.7$^{\blacktriangledown}$ & 51.8$^{\blacktriangledown}$ & 66.2$^{\blacktriangledown}$ & \textbf{44.4} & 105.4$^{\blacktriangledown}$ & 78.4$^{\blacktriangledown}$ & 68.6$^{\blacktriangledown}$ & 74.5$^{\blacktriangledown}$ \\
 &  & Nomic Vision & \textbf{39.3} & 95.0$^{\blacktriangledown}$ & 74.2$^{\blacktriangledown}$ & 60.7$^{\blacktriangledown}$ & 79.1$^{\blacktriangledown}$ & \textbf{46.4} & 113.5$^{\blacktriangledown}$ & 88.9$^{\blacktriangledown}$ & 77.7$^{\blacktriangledown}$ & 92.4$^{\blacktriangledown}$ \\
\cmidrule(lr){2-13}
 & Mistral-Small-3.2 & PE-Core & \textbf{90.9} & 157.6$^{\blacktriangledown}$ & 139.2$^{\blacktriangledown}$ & 131.0$^{\blacktriangledown}$ & 131.0$^{\blacktriangledown}$ & \textbf{99.3} & 157.9$^{\blacktriangledown}$ & 141.1$^{\blacktriangledown}$ & 149.6$^{\blacktriangledown}$ & 145.9$^{\blacktriangledown}$ \\
 &  & SigLIP2 & \textbf{126.5} & 217.4$^{\blacktriangledown}$ & 219.3$^{\blacktriangledown}$ & 172.2$^{\blacktriangledown}$ & 174.6$^{\blacktriangledown}$ & \textbf{133.7} & 220.7$^{\blacktriangledown}$ & 219.0$^{\blacktriangledown}$ & 191.4$^{\blacktriangledown}$ & 182.6$^{\blacktriangledown}$ \\
 &  & Nomic Vision & \textbf{108.0} & 154.9$^{\blacktriangledown}$ & 175.6$^{\blacktriangledown}$ & 148.8$^{\blacktriangledown}$ & 144.3 & \textbf{113.2} & 163.5$^{\blacktriangledown}$ & 180.7$^{\blacktriangledown}$ & 169.2$^{\blacktriangledown}$ & 159.1$^{\blacktriangledown}$ \\
\cmidrule(lr){2-13}
 & GLM-5.3-Flash & PE-Core & \textbf{32.5} & 44.6$^{\blacktriangledown}$ & 47.9$^{\blacktriangledown}$ & 47.8$^{\blacktriangledown}$ & 48.4$^{\blacktriangledown}$ & \textbf{38.0} & 62.5$^{\blacktriangledown}$ & 59.0$^{\blacktriangledown}$ & 64.9$^{\blacktriangledown}$ & 57.8$^{\blacktriangledown}$ \\
 &  & SigLIP2 & \textbf{35.8} & 49.9 & 52.2$^{\blacktriangledown}$ & 50.2$^{\blacktriangledown}$ & 53.6$^{\blacktriangledown}$ & \textbf{41.9} & 62.8$^{\blacktriangledown}$ & 66.7$^{\blacktriangledown}$ & 66.1$^{\blacktriangledown}$ & 62.3$^{\blacktriangledown}$ \\
 &  & Nomic Vision & \textbf{35.5} & 49.8$^{\blacktriangledown}$ & 60.8$^{\blacktriangledown}$ & 56.0$^{\blacktriangledown}$ & 59.7$^{\blacktriangledown}$ & \textbf{40.2} & 63.6$^{\blacktriangledown}$ & 72.9$^{\blacktriangledown}$ & 71.2$^{\blacktriangledown}$ & 65.6$^{\blacktriangledown}$ \\
\midrule
MADiMa & GPT-5.4-mini & PE-Core & \textbf{32.3} & 121.4$^{\blacktriangledown}$ & 109.8$^{\blacktriangledown}$ & 71.6$^{\blacktriangledown}$ & 77.7$^{\blacktriangledown}$ & \textbf{42.1} & 129.0$^{\blacktriangledown}$ & 122.0$^{\blacktriangledown}$ & 94.0$^{\blacktriangledown}$ & 82.7$^{\blacktriangledown}$ \\
 &  & SigLIP2 & \textbf{32.0} & 140.5$^{\blacktriangledown}$ & 111.5$^{\blacktriangledown}$ & 72.9$^{\blacktriangledown}$ & 85.7$^{\blacktriangledown}$ & \textbf{43.1} & 133.0$^{\blacktriangledown}$ & 128.8$^{\blacktriangledown}$ & 100.5$^{\blacktriangledown}$ & 85.4$^{\blacktriangledown}$ \\
 &  & Nomic Vision & \textbf{33.2} & 114.5$^{\blacktriangledown}$ & 119.4$^{\blacktriangledown}$ & 75.7$^{\blacktriangledown}$ & 90.8$^{\blacktriangledown}$ & \textbf{42.3} & 115.9$^{\blacktriangledown}$ & 132.1$^{\blacktriangledown}$ & 104.0$^{\blacktriangledown}$ & 92.0$^{\blacktriangledown}$ \\
\cmidrule(lr){2-13}
 & Gemma-4-31B & PE-Core & \textbf{36.7} & 157.6$^{\blacktriangledown}$ & 125.2$^{\blacktriangledown}$ & 82.2$^{\blacktriangledown}$ & 91.1$^{\blacktriangledown}$ & \textbf{46.2} & 147.6$^{\blacktriangledown}$ & 128.5$^{\blacktriangledown}$ & 102.5$^{\blacktriangledown}$ & 90.1$^{\blacktriangledown}$ \\
 &  & SigLIP2 & \textbf{34.8} & 138.2$^{\blacktriangledown}$ & 118.5$^{\blacktriangledown}$ & 75.7$^{\blacktriangledown}$ & 94.2$^{\blacktriangledown}$ & \textbf{42.5} & 135.4$^{\blacktriangledown}$ & 124.4$^{\blacktriangledown}$ & 98.9$^{\blacktriangledown}$ & 93.8$^{\blacktriangledown}$ \\
 &  & Nomic Vision & \textbf{37.0} & 158.2$^{\blacktriangledown}$ & 118.3$^{\blacktriangledown}$ & 82.3$^{\blacktriangledown}$ & 90.7$^{\blacktriangledown}$ & \textbf{45.1} & 150.8$^{\blacktriangledown}$ & 125.2$^{\blacktriangledown}$ & 104.4$^{\blacktriangledown}$ & 91.6$^{\blacktriangledown}$ \\
\cmidrule(lr){2-13}
 & Mistral-Small-3.2 & PE-Core & \textbf{48.6} & 130.3$^{\blacktriangledown}$ & 109.9$^{\blacktriangledown}$ & 72.6$^{\blacktriangledown}$ & 98.3$^{\blacktriangledown}$ & \textbf{63.5} & 143.9$^{\blacktriangledown}$ & 114.4$^{\blacktriangledown}$ & 87.5$^{\blacktriangledown}$ & 101.3$^{\blacktriangledown}$ \\
 &  & SigLIP2 & \textbf{59.8} & 118.4$^{\blacktriangledown}$ & 135.6$^{\blacktriangledown}$ & 85.7$^{\blacktriangledown}$ & 95.0$^{\blacktriangledown}$ & \textbf{74.1} & 120.1$^{\blacktriangledown}$ & 137.0$^{\blacktriangledown}$ & 105.8$^{\blacktriangledown}$ & 98.6$^{\blacktriangledown}$ \\
 &  & Nomic Vision & \textbf{53.6} & 93.2$^{\blacktriangledown}$ & 115.4$^{\blacktriangledown}$ & 83.5$^{\blacktriangledown}$ & 90.1$^{\blacktriangledown}$ & \textbf{67.5} & 105.7$^{\blacktriangledown}$ & 128.0$^{\blacktriangledown}$ & 111.5$^{\blacktriangledown}$ & 102.7$^{\blacktriangledown}$ \\
\cmidrule(lr){2-13}
 & GLM-5.3-Flash & PE-Core & \textbf{30.4} & 62.0$^{\blacktriangledown}$ & 94.9$^{\blacktriangledown}$ & 50.8$^{\blacktriangledown}$ & 48.8$^{\blacktriangledown}$ & \textbf{37.3} & 69.6$^{\blacktriangledown}$ & 97.5$^{\blacktriangledown}$ & 66.4$^{\blacktriangledown}$ & 65.8$^{\blacktriangledown}$ \\
 &  & SigLIP2 & \textbf{29.5} & 82.0$^{\blacktriangledown}$ & 93.4$^{\blacktriangledown}$ & 54.7$^{\blacktriangledown}$ & 57.6$^{\blacktriangledown}$ & \textbf{38.4} & 85.1$^{\blacktriangledown}$ & 98.8$^{\blacktriangledown}$ & 71.5$^{\blacktriangledown}$ & 75.1$^{\blacktriangledown}$ \\
 &  & Nomic Vision & \textbf{31.7} & 66.4$^{\blacktriangledown}$ & 95.2$^{\blacktriangledown}$ & 50.5$^{\blacktriangledown}$ & 44.7$^{\blacktriangledown}$ & \textbf{37.6} & 70.8$^{\blacktriangledown}$ & 97.8$^{\blacktriangledown}$ & 67.8$^{\blacktriangledown}$ & 61.5$^{\blacktriangledown}$ \\
\midrule
EDUCATE & GPT-5.4-mini & PE-Core & -- & -- & -- & -- & -- & \textbf{36.1} & 108.4$^{\blacktriangledown}$ & 45.7$^{\blacktriangledown}$ & 105.7$^{\blacktriangledown}$ & 103.3$^{\blacktriangledown}$ \\
 &  & SigLIP2 & -- & -- & -- & -- & -- & \textbf{38.0} & 98.1$^{\blacktriangledown}$ & 45.0 & 119.1$^{\blacktriangledown}$ & 110.7$^{\blacktriangledown}$ \\
 &  & Nomic Vision & -- & -- & -- & -- & -- & \textbf{41.9} & 105.8$^{\blacktriangledown}$ & 64.4$^{\blacktriangledown}$ & 108.0$^{\blacktriangledown}$ & 108.4$^{\blacktriangledown}$ \\
\cmidrule(lr){2-13}
 & Gemma-4-31B & PE-Core & -- & -- & -- & -- & -- & \textbf{38.5} & 99.4$^{\blacktriangledown}$ & 46.9 & 107.8$^{\blacktriangledown}$ & 109.8$^{\blacktriangledown}$ \\
 &  & SigLIP2 & -- & -- & -- & -- & -- & \textbf{41.9} & 87.3$^{\blacktriangledown}$ & 47.2 & 108.4$^{\blacktriangledown}$ & 99.3$^{\blacktriangledown}$ \\
 &  & Nomic Vision & -- & -- & -- & -- & -- & \textbf{40.4} & 97.6$^{\blacktriangledown}$ & 46.9 & 116.2$^{\blacktriangledown}$ & 113.9$^{\blacktriangledown}$ \\
\cmidrule(lr){2-13}
 & Mistral-Small-3.2 & PE-Core & -- & -- & -- & -- & -- & \textbf{118.0} & 133.4 & 123.5 & 165.2$^{\blacktriangledown}$ & 188.8 \\
 &  & SigLIP2 & -- & -- & -- & -- & -- & \textbf{134.5} & 139.6 & 156.7 & 168.4$^{\blacktriangledown}$ & 161.9 \\
 &  & Nomic Vision & -- & -- & -- & -- & -- & \textbf{140.3} & 153.7 & 152.8 & 184.0$^{\blacktriangledown}$ & 169.9 \\
\cmidrule(lr){2-13}
 & GLM-5.3-Flash & PE-Core & -- & -- & -- & -- & -- & \textbf{51.9} & 137.9$^{\blacktriangledown}$ & 61.6 & 92.1$^{\blacktriangledown}$ & 81.1$^{\blacktriangledown}$ \\
 &  & SigLIP2 & -- & -- & -- & -- & -- & \textbf{52.5} & 123.8$^{\blacktriangledown}$ & 57.2 & 77.6 & 76.4$^{\blacktriangledown}$ \\
 &  & Nomic Vision & -- & -- & -- & -- & -- & \textbf{55.1} & 147.6$^{\blacktriangledown}$ & 58.5 & 94.9$^{\blacktriangledown}$ & 88.3$^{\blacktriangledown}$ \\
\bottomrule
\end{tabular}%
}
\end{table}

Table~\ref{tab:post-grounding} reports the resulting $\mathrm{\%MAE}_{\text{energy}}$ and $\mathrm{\%MAE}_{\text{micro/CHO}}$.
The analysis covers 48 model--encoder--query-set configurations (4 VLMs $\times$ 3 encoders $\times$ 4 query sets).
The original NutriVisRAG estimate achieves the lowest $\mathrm{\%MAE}_{\text{micro/CHO}}$ in 46/48 configurations and the lowest $\mathrm{\%MAE}_{\text{energy}}$ in 34/36 configurations where energy is available.
Lexical, EmbeddingGemma, and Nomic Text matching increase $\mathrm{\%MAE}_{\text{micro/CHO}}$ in all 48 configurations, while Qwen3 increases it in 46/48.
After Benjamini--Hochberg correction, significant degradations occur in 45/48 configurations for lexical matching, 30/48 for Qwen3, 47/48 for EmbeddingGemma, and 42/48 for Nomic Text, with no significant improvement for any matching method.
Because food identification, database-match correctness, and item-weight estimation are not evaluated independently, these findings characterise the end-to-end post-hoc pipeline rather than the matching models alone.
Post-hoc food-composition grounding therefore provides no statistically significant improvement over the original NutriVisRAG predictions.
Furthermore, it adds little computational overhead: embedding-based matching takes about 150\,ms per predicted food item, while lexical matching takes about 1.5\,ms.
Its exclusion is therefore driven by degraded accuracy rather than computational cost.

\subsection{Retrieval Reranking}
\label{sec:reranking}

We additionally investigate whether reranking a larger pool of visually retrieved candidates improves nutrient estimation.
Rather than directly presenting the five nearest neighbours returned by the visual encoder, we retrieve a larger candidate pool and apply a multimodal relevance model before selecting the final exemplars provided to the VLM.

\paragraph{Reranking strategies.}
For each query, the visual encoder first retrieves 50 candidate reference meals.
We evaluate three multimodal rerankers: Jina Reranker M0~\cite{jina2025rerankerm0}, Qwen3-VL-Reranker-2B~\cite{li2026qwen3vlreranker}, and Llama Nemotron Rerank VL 1B v2~\cite{nvidia2025nemotronrerankvl}.
Jina and Qwen3 perform image-to-image reranking by scoring the query meal image against each candidate image.
Nemotron performs text-to-image reranking, for which the query meal is first converted into a short food and portion description using the same VLM used for nutrient estimation.
Each reranker rescores the 50 candidates and retains the five highest-ranked references.

\paragraph{Experimental setup.}
The analysis uses the same fixed 100-query subsets as the retrieval ablations and evaluates both per-item and whole-meal prompting.
The evaluated checkpoints are jinaai/jina-reranker-m0, Qwen/Qwen3-VL-Reranker-2B, and nvidia/llama-nemotron-rerank-vl-1b-v2.
The initial candidate pool contains 50 references, which are reranked to the five exemplars presented to the VLM.
All remaining retrieval and inference settings are kept unchanged from the corresponding NutriVisRAG configuration.
The default NutriVisRAG result without reranking serves as the baseline for each model--encoder--query-set--prompt combination.
The rerankers are run locally on an NVIDIA RTX A6000.

Table~\ref{tab:rerank} reports the resulting $\mathrm{\%MAE}_{\text{energy}}$ and $\mathrm{\%MAE}_{\text{micro/CHO}}$.

\begin{table*}[tb]
\centering
\caption{Multimodal reranking on the fixed 100-query subsets, with per-item and whole-meal prompting side by side (\%MAE, lower is better); 50 candidates are reranked to five exemplars.
NutriVisRAG denotes the default configuration without reranking.
Dashes indicate unavailable evaluation targets for the corresponding dataset.
Bold indicates the lowest error per row and metric; $^\star$/$^{\blacktriangledown}$ indicate significant improvements/degradations relative to NutriVisRAG (paired Wilcoxon signed-rank test, Benjamini--Hochberg corrected, $q<0.05$).
}
\label{tab:rerank}
\scriptsize
\setlength{\tabcolsep}{1.6pt}
\renewcommand{\arraystretch}{0.95}
\resizebox{\textwidth}{!}{%
\begin{tabular}{lll rrrr rrrr rrrr rrrr}
\toprule
& & & \multicolumn{8}{c}{\textbf{Per-item}} & \multicolumn{8}{c}{\textbf{Whole-meal}} \\
\cmidrule(lr){4-11}\cmidrule(lr){12-19}
& & & \multicolumn{4}{c}{\textbf{Energy}} & \multicolumn{4}{c}{\textbf{Micro / CHO}} & \multicolumn{4}{c}{\textbf{Energy}} & \multicolumn{4}{c}{\textbf{Micro / CHO}} \\
\cmidrule(lr){4-7}\cmidrule(lr){8-11}\cmidrule(lr){12-15}\cmidrule(lr){16-19}
\textbf{Query set} & \textbf{Model} & \textbf{Encoder} & \textbf{Default} & \textbf{Jina} & \textbf{Qwen3} & \textbf{Nem.} & \textbf{Default} & \textbf{Jina} & \textbf{Qwen3} & \textbf{Nem.} & \textbf{Default} & \textbf{Jina} & \textbf{Qwen3} & \textbf{Nem.} & \textbf{Default} & \textbf{Jina} & \textbf{Qwen3} & \textbf{Nem.} \\
\midrule
Nutrition5k & GPT-5.4-mini & PE-Core & 32.8 & 29.9 & \textbf{28.7} & 38.5 & 40.0 & \textbf{35.2} & 35.6 & 43.5 & 34.7 & \textbf{31.5} & 36.8 & 47.4 & 40.4 & \textbf{36.9} & 40.9 & 49.8 \\
 &  & SigLIP2 & 30.2 & 33.0 & 30.6 & \textbf{30.2} & 35.6 & 41.7 & \textbf{34.2} & 36.7 & 36.6 & \textbf{33.1} & 34.6 & 37.6 & 39.9 & \textbf{39.7} & 40.5 & 41.6 \\
 &  & Nomic Vision & 29.9 & 30.5 & \textbf{27.6} & 37.6 & 37.6 & 38.2 & \textbf{36.6} & 45.1 & 40.1 & 38.5 & \textbf{36.6} & 39.6 & 46.8 & 44.1 & \textbf{40.3} & 47.6 \\
\cmidrule(lr){2-19}
 & Gemma-4-31B & PE-Core & \textbf{50.1} & 57.1 & 51.5 & 71.9$^{\blacktriangledown}$ & 60.4 & 63.9 & \textbf{55.3} & 77.8$^{\blacktriangledown}$ & \textbf{44.0} & 50.4 & 50.5 & 65.5$^{\blacktriangledown}$ & \textbf{50.0} & 57.1 & 57.3 & 68.5$^{\blacktriangledown}$ \\
 &  & SigLIP2 & \textbf{44.8} & 65.7 & 60.4 & 72.9$^{\blacktriangledown}$ & \textbf{49.8} & 71.3 & 65.9 & 79.2$^{\blacktriangledown}$ & \textbf{44.3} & 50.4 & 55.6 & 69.7 & \textbf{48.4} & 60.6 & 60.7 & 75.6$^{\blacktriangledown}$ \\
 &  & Nomic Vision & 72.2 & 77.7 & \textbf{67.8} & 69.9 & 80.7 & 84.9 & \textbf{76.2} & 77.3 & \textbf{55.7} & 66.9 & 56.2 & 57.6 & 62.5 & 71.1 & \textbf{61.0} & 62.4 \\
\cmidrule(lr){2-19}
 & Mistral-Small-3.2 & PE-Core & 244.8 & \textbf{89.0}$^\star$ & 109.1$^\star$ & 186.7 & 259.9 & \textbf{107.3}$^\star$ & 126.9$^\star$ & 200.6 & \textbf{70.0} & 79.2 & 85.4 & 88.5 & \textbf{82.3} & 89.5 & 94.0 & 100.9 \\
 &  & SigLIP2 & 283.1 & \textbf{94.7}$^\star$ & 136.5$^\star$ & 231.6 & 306.5 & \textbf{108.5}$^\star$ & 147.2$^\star$ & 224.1 & \textbf{78.5} & 84.0 & 88.0 & 107.5 & \textbf{86.6} & 93.1 & 101.2 & 119.7 \\
 &  & Nomic Vision & 418.6 & \textbf{154.3}$^\star$ & 158.1$^\star$ & 233.3$^\star$ & 410.9 & 174.1$^\star$ & \textbf{169.0}$^\star$ & 237.4$^\star$ & 117.6 & 108.9 & \textbf{92.5} & 118.4 & 124.0 & 110.2 & \textbf{109.7} & 127.3 \\
\cmidrule(lr){2-19}
 & GLM-5.3-Flash & PE-Core & 50.8 & \textbf{38.1} & 41.1 & 53.7 & 52.5 & \textbf{41.4} & 42.9 & 53.9 & 38.4 & \textbf{36.5} & 44.3 & 53.2 & 39.8 & \textbf{39.1} & 44.7 & 54.9 \\
 &  & SigLIP2 & 36.5 & 42.5 & \textbf{36.3} & 55.4 & \textbf{40.7} & 44.8 & 40.8 & 57.4 & \textbf{30.5} & 38.1 & 31.3 & 46.9 & \textbf{33.5} & 40.5 & 35.6 & 51.8$^{\blacktriangledown}$ \\
 &  & Nomic Vision & 36.3 & 45.1 & \textbf{35.3} & 39.0 & 42.0 & 48.0 & \textbf{37.5} & 44.8 & 36.2 & \textbf{33.3} & 39.0 & 38.2 & 40.9 & \textbf{37.8} & 41.6 & 41.0 \\
\midrule
SNAPMe & GPT-5.4-mini & PE-Core & \textbf{34.5} & 37.4 & 35.3 & 37.7 & \textbf{42.0} & 42.9 & 43.3 & 42.8 & \textbf{33.0} & 39.1 & 37.5 & 34.5 & \textbf{41.3} & 43.8 & 44.0 & 42.0 \\
 &  & SigLIP2 & \textbf{34.5} & 34.6 & 36.2 & 35.2 & 42.4 & 42.2 & 42.7 & \textbf{42.0} & \textbf{33.6} & 34.9 & 36.2 & 36.9 & \textbf{41.5} & 41.6 & 42.4 & 43.7 \\
 &  & Nomic Vision & 36.1 & \textbf{34.2} & 36.0 & 34.9 & 42.4 & \textbf{42.3} & 43.5 & 42.6 & 36.3 & \textbf{35.4} & 36.6 & 38.6 & 43.1 & \textbf{42.7} & 43.9 & 44.8 \\
\cmidrule(lr){2-19}
 & Gemma-4-31B & PE-Core & \textbf{38.8} & 39.3 & 42.9 & 42.9 & \textbf{44.2} & 45.6 & 48.2 & 46.5 & 38.1 & 38.6 & 38.6 & \textbf{37.9} & 42.5 & 41.6 & 43.4 & \textbf{41.4} \\
 &  & SigLIP2 & 37.9 & 45.8 & \textbf{36.2} & 42.4 & 44.4 & 51.1 & \textbf{43.0} & 49.5 & 36.8 & 36.9 & 37.1 & \textbf{36.6} & 41.2 & 41.1 & \textbf{40.8} & 41.0 \\
 &  & Nomic Vision & 39.3 & 37.3 & \textbf{35.8} & 37.2 & 46.4 & 42.7 & \textbf{42.6} & 44.9 & 36.3 & \textbf{34.8} & 37.1 & 35.3 & 41.5 & \textbf{40.2} & 41.6 & 41.4 \\
\cmidrule(lr){2-19}
 & Mistral-Small-3.2 & PE-Core & 90.9 & 120.2 & \textbf{90.6} & 98.3 & 99.3 & 123.6 & \textbf{95.3} & 105.1 & 57.5 & 59.8 & \textbf{53.2} & 59.4 & 66.6 & 66.5 & \textbf{58.9} & 66.8 \\
 &  & SigLIP2 & 126.5 & 116.8 & \textbf{113.9} & 142.8 & 133.7 & 125.2 & \textbf{123.0} & 146.3 & 66.3 & 60.2 & \textbf{54.2} & 57.5 & 76.0 & 70.9 & \textbf{65.9} & 66.4 \\
 &  & Nomic Vision & 108.0 & 138.5 & \textbf{95.9} & 110.5 & 113.2 & 145.4 & \textbf{100.0} & 115.1 & 59.5 & 61.7 & 51.7 & \textbf{49.2} & 67.4 & 69.5 & \textbf{60.3} & 61.1 \\
\cmidrule(lr){2-19}
 & GLM-5.3-Flash & PE-Core & \textbf{32.5} & 34.9 & 32.8 & 34.2 & \textbf{38.0} & 40.6 & 38.5 & 39.1 & \textbf{30.8} & 32.8 & 34.7 & 34.2 & \textbf{35.8} & 38.0 & 38.7 & 39.0 \\
 &  & SigLIP2 & 35.8 & 34.8 & 39.4 & \textbf{32.8} & 41.9 & 39.8 & 44.5 & \textbf{38.8} & 33.7 & 33.4 & \textbf{33.0} & 38.4 & 39.0 & \textbf{38.4} & 39.0 & 43.4 \\
 &  & Nomic Vision & \textbf{35.5} & \textbf{35.5} & 44.1 & 39.5 & \textbf{40.2} & 40.9 & 47.1 & 43.9 & \textbf{32.4} & 36.3 & 33.2 & 33.6 & 38.5 & 40.5 & \textbf{37.4} & 38.6 \\
\midrule
MADiMa & GPT-5.4-mini & PE-Core & 32.3 & \textbf{31.2} & 33.4 & 32.7 & 42.1 & \textbf{41.7} & 41.7 & 42.8 & 37.6 & 37.0 & 39.2 & \textbf{33.9} & 44.6 & 44.9 & 45.5 & \textbf{42.4} \\
 &  & SigLIP2 & 32.0 & \textbf{30.4} & 31.0 & 33.5 & 43.1 & \textbf{40.5} & 40.7 & 43.4 & 45.2 & 38.4 & \textbf{38.1} & 38.6 & 53.0 & \textbf{46.3} & 46.7 & 49.1 \\
 &  & Nomic Vision & 33.2 & 32.4 & 35.2 & \textbf{31.9} & 42.3 & \textbf{40.5} & 42.8 & 42.1 & 41.8 & 37.9 & 41.9 & \textbf{36.9} & 47.9 & 46.0 & 48.1 & \textbf{44.8} \\
\cmidrule(lr){2-19}
 & Gemma-4-31B & PE-Core & 36.7 & 38.6 & 35.8 & \textbf{35.7} & 46.2 & 46.2 & \textbf{44.4} & 44.8 & 57.4 & 54.0 & 53.3 & \textbf{53.0} & 59.3 & 56.7 & 55.5 & \textbf{55.2} \\
 &  & SigLIP2 & \textbf{34.8} & 38.4 & 35.2 & 36.6 & 42.5 & 46.1 & \textbf{42.4} & 46.3 & 56.6 & 55.9 & 51.4 & \textbf{51.4}$^\star$ & 58.2 & 57.7 & 53.8 & \textbf{53.6}$^\star$ \\
 &  & Nomic Vision & 37.0 & 35.0 & \textbf{34.8} & 37.2 & 45.1 & 43.8 & \textbf{41.6} & 45.2 & 52.2 & 47.4 & \textbf{45.1} & 48.5 & 55.3 & 52.2 & \textbf{51.6} & 53.4 \\
\cmidrule(lr){2-19}
 & Mistral-Small-3.2 & PE-Core & 48.6 & \textbf{42.8} & 48.7 & 44.2 & 63.5 & 57.8 & 59.1 & \textbf{55.3} & 40.7 & 40.8 & 41.8 & \textbf{36.4} & 49.5 & 49.9 & 54.0 & \textbf{49.2} \\
 &  & SigLIP2 & 59.8 & 47.5 & 48.7 & \textbf{45.5} & 74.1 & 58.7 & 59.3 & \textbf{52.8} & \textbf{38.7} & 39.4 & 41.1 & 42.1 & 54.0 & 50.2 & 52.2 & \textbf{49.8} \\
 &  & Nomic Vision & 53.6 & \textbf{48.0} & 58.1 & 50.8 & 67.5 & \textbf{59.7} & 72.9 & 61.3 & 50.7 & \textbf{37.5} & 42.8 & 40.5 & 66.0 & \textbf{46.4} & 55.0 & 51.4 \\
\cmidrule(lr){2-19}
 & GLM-5.3-Flash & PE-Core & 30.4 & 30.2 & \textbf{29.4} & 30.0 & 37.3 & 38.7 & 38.1 & \textbf{37.1} & \textbf{29.8} & 31.1 & 30.2 & 31.2 & 37.9 & \textbf{37.0} & 38.1 & 37.7 \\
 &  & SigLIP2 & 29.5 & 31.1 & \textbf{27.4} & 29.4 & 38.4 & 38.1 & 36.4 & \textbf{36.2} & 29.6 & 29.8 & \textbf{27.4} & 29.8 & 36.9 & \textbf{35.4} & 36.7 & 36.9 \\
 &  & Nomic Vision & 31.7 & 32.8 & \textbf{28.7} & 31.5 & 37.6 & 41.4 & \textbf{36.4} & 38.3 & 31.2 & 30.4 & \textbf{29.2} & 29.4 & 37.5 & 37.5 & 37.0 & \textbf{36.7} \\
\midrule
EDUCATE & GPT-5.4-mini & PE-Core & -- & -- & -- & -- & \textbf{36.1} & 46.0 & 36.4 & 43.8 & -- & -- & -- & -- & \textbf{41.1} & 42.4 & 49.2 & 41.5 \\
 &  & SigLIP2 & -- & -- & -- & -- & 38.0 & 42.6 & 40.1 & \textbf{36.5} & -- & -- & -- & -- & \textbf{41.6} & 47.5 & 43.9 & 45.0 \\
 &  & Nomic Vision & -- & -- & -- & -- & 41.9 & 42.9 & \textbf{40.8} & 44.3 & -- & -- & -- & -- & \textbf{42.9} & 43.3 & 53.7 & 49.2 \\
\cmidrule(lr){2-19}
 & Gemma-4-31B & PE-Core & -- & -- & -- & -- & 38.5 & \textbf{32.1} & 38.0 & 34.4 & -- & -- & -- & -- & 46.8 & \textbf{39.5} & 45.4 & 42.0 \\
 &  & SigLIP2 & -- & -- & -- & -- & 41.9 & \textbf{29.8}$^\star$ & 39.4 & 37.4 & -- & -- & -- & -- & 48.3 & \textbf{33.0}$^\star$ & 47.7 & 45.5 \\
 &  & Nomic Vision & -- & -- & -- & -- & 40.4 & 40.7 & 41.0 & \textbf{39.0} & -- & -- & -- & -- & 50.5 & \textbf{37.8}$^\star$ & 47.1 & 41.5 \\
\cmidrule(lr){2-19}
 & Mistral-Small-3.2 & PE-Core & -- & -- & -- & -- & 118.0 & 110.8 & \textbf{98.2} & 120.5 & -- & -- & -- & -- & 91.0 & 56.5 & 65.9 & \textbf{55.9} \\
 &  & SigLIP2 & -- & -- & -- & -- & 134.5 & 116.7 & 119.3 & \textbf{115.1} & -- & -- & -- & -- & 96.8 & 62.2$^\star$ & 58.9$^\star$ & \textbf{58.4}$^\star$ \\
 &  & Nomic Vision & -- & -- & -- & -- & 140.3 & \textbf{111.7} & 126.8 & 147.8 & -- & -- & -- & -- & 90.7 & \textbf{60.0} & 66.1 & 77.5 \\
\cmidrule(lr){2-19}
 & GLM-5.3-Flash & PE-Core & -- & -- & -- & -- & \textbf{51.9} & 57.2 & 53.9 & 56.8 & -- & -- & -- & -- & 46.7 & \textbf{41.5} & 43.0 & 56.3$^{\blacktriangledown}$ \\
 &  & SigLIP2 & -- & -- & -- & -- & \textbf{52.5} & 61.4 & 55.1 & 66.0 & -- & -- & -- & -- & \textbf{46.2} & 49.5 & 47.1 & 54.7 \\
 &  & Nomic Vision & -- & -- & -- & -- & 55.1 & 55.1 & 56.6 & \textbf{51.5} & -- & -- & -- & -- & 44.0 & \textbf{43.5} & 49.5 & 49.0 \\
\bottomrule
\end{tabular}%
}
\end{table*}

Under per-item prompting, reranking reduces $\mathrm{\%MAE}_{\text{micro/CHO}}$ in 73/144 comparisons, with eight significant improvements and two significant degradations after Benjamini--Hochberg correction.
Improvements occur in 24/36 comparisons on MADiMa, 19/36 on Nutrition5k, 17/36 on EDUCATE, and 13/36 on SNAPMe.
Jina improves 23/48 per-item comparisons, Qwen3 improves 30/48, and Nemotron improves 20/48.
Under whole-meal prompting, reranking reduces $\mathrm{\%MAE}_{\text{micro/CHO}}$ in 76/144 comparisons, with six significant improvements and four significant degradations after correction.
Improvements occur in 28/36 comparisons on MADiMa, 21/36 on EDUCATE, 17/36 on SNAPMe, and 10/36 on Nutrition5k.
Jina improves 30/48 whole-meal comparisons, while Qwen3 and Nemotron each improve 23/48.
These results indicate that reranking can reduce nutrient-estimation error in certain configurations, but its effects depend on the dataset, prompting strategy, VLM, and reranker.
Reranking adds a median of 5.3--11.7\,s per meal, increasing end-to-end latency by $2.6\times$--$5.1\times$, depending on the reranker and VLM.
Given the variable accuracy gains and additional computational cost, reranking is retained as an exploratory extension rather than incorporated into the default NutriVisRAG pipeline.

\end{document}